\documentclass[letterpaper, 10 pt, conference]{ieeeconf}

\IEEEoverridecommandlockouts
\usepackage{amsmath,amsfonts}
\usepackage{array}
\usepackage{booktabs}
\usepackage{multirow}
\usepackage{caption}
\usepackage{graphicx}
\usepackage{epstopdf}
\usepackage{float}
\usepackage{subfigure}
\graphicspath{{./figure_MVSEC/}}
\makeatletter
\renewcommand{\@thesubfigure}{\hskip\subfiglabelskip}
\makeatother

\usepackage{hyperref}
\hypersetup{hypertex=true,
colorlinks=true,
linkcolor=red,
anchorcolor=blue,
citecolor=green}
\usepackage{cite}
\usepackage{xcolor}
\usepackage{hyperref}

\title{\LARGE \bf
Nonlinear Motion-Guided and Spatio-Temporal Aware Network for Unsupervised Event-Based Optical Flow
}

\author{Zuntao Liu, Hao Zhuang, Junjie Jiang, Yuhang Song and Zheng Fang*%
\thanks{This work was supported by the National Natural Science Foundation of China under Grants 62073066, the Fundamental Research Funds for Central Universities under Grant N2226001, and 111 Project under Grant B16009. (Corresponding author: Zheng Fang, e-mail: fangzheng@mail.neu.edu.cn)}%
\thanks{The authors are all with the Faculty of Robot Science and Engineering, Northeastern University, Shenyang, China;}
}

\begin{document}
\bstctlcite{IEEEexample:BSTcontrol}

\maketitle
\thispagestyle{empty}
\pagestyle{empty}

\begin{abstract}
Event cameras have the potential to capture continuous motion information over time and space, making them well-suited for optical flow estimation. However, most existing learning-based methods for event-based optical flow adopt frame-based techniques, ignoring the spatio-temporal characteristics of events.
Additionally, these methods assume linear motion between consecutive events within the loss time window, which increases optical flow errors in long-time sequences. 
In this work, we observe that rich spatio-temporal information and accurate nonlinear motion between events are crucial for event-based optical flow estimation. Therefore, we propose E-NMSTFlow, a novel unsupervised event-based optical flow network focusing on long-time sequences. We propose a Spatio-Temporal Motion Feature Aware (STMFA) module and an Adaptive Motion Feature Enhancement (AMFE) module, both of which utilize rich spatio-temporal information to learn spatio-temporal data associations. Meanwhile, we propose a nonlinear motion compensation loss that utilizes the accurate nonlinear motion between events to improve the unsupervised learning of our network. Extensive experiments demonstrate the effectiveness and superiority of our method. Remarkably, our method ranks first among unsupervised learning methods on the MVSEC and DSEC-Flow datasets. Our project page is available at \href{https://wynelio.github.io/E-NMSTFlow}{\textcolor{magenta}{https://wynelio.github.io/E-NMSTFlow}}.
\end{abstract}

\section{Introduction}
Optical flow estimation is a fundamental task in robotic vision, which aims to calculate the motion vector of each pixel on the image plane without geometric prior. Owing to the ability to encode primitive motion information, it plays an important role in many robotic applications, such as visual odometry\cite{qin2018vins,vidal2018ultimate}, robot navigation\cite{chao2014survey,bouwmeester2023nanoflownet,jiang2023neuro}, autonomous driving\cite{geiger2012we}, and object tracking\cite{piga2021roft,zheng2022spike}. 
\begin{figure}[t]
\subfigure[(a) Image]{
    \begin{minipage}[b]{0.285\linewidth}
        \centering
        \includegraphics[scale=0.166]{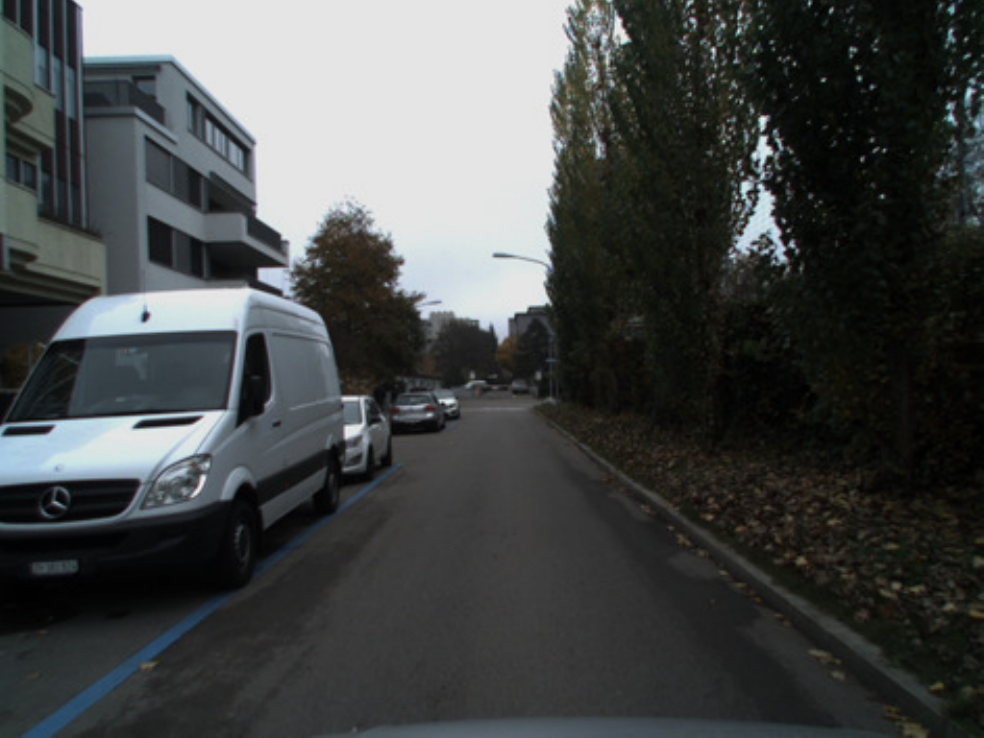}\\
        \vspace{1.6pt}
        \includegraphics[scale=0.166]{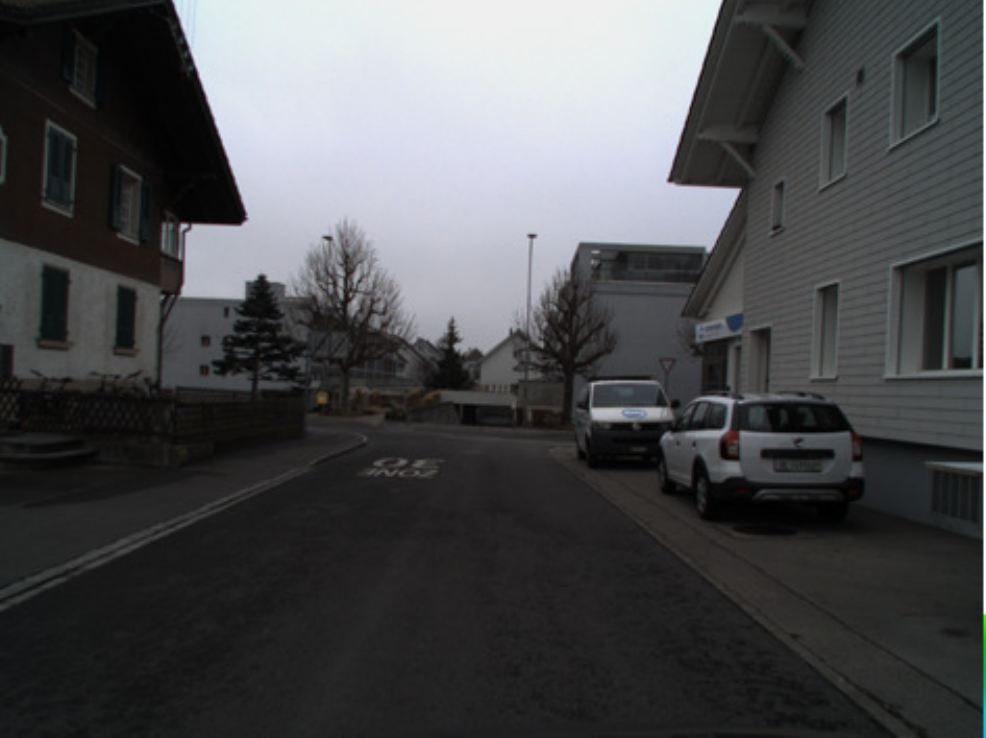}\\
    \end{minipage}
    }
\subfigure[(b) Events]{
    \begin{minipage}[b]{0.285\linewidth}
        \centering
     \includegraphics[scale=0.166]{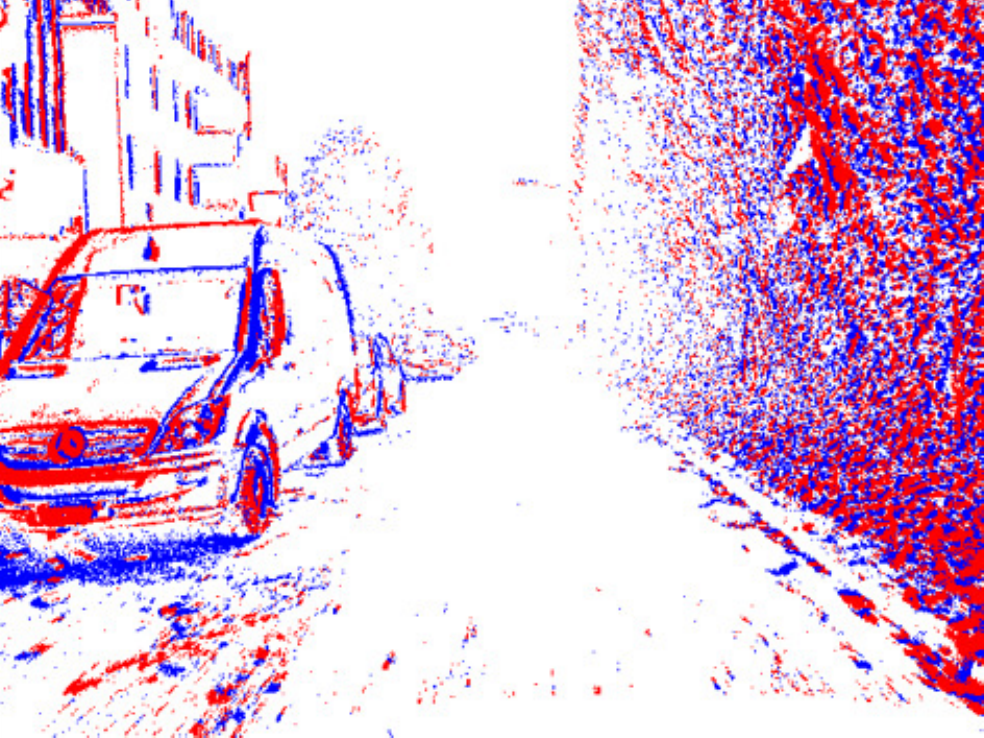}\\
        \vspace{1.6pt}
        \includegraphics[scale=0.166]{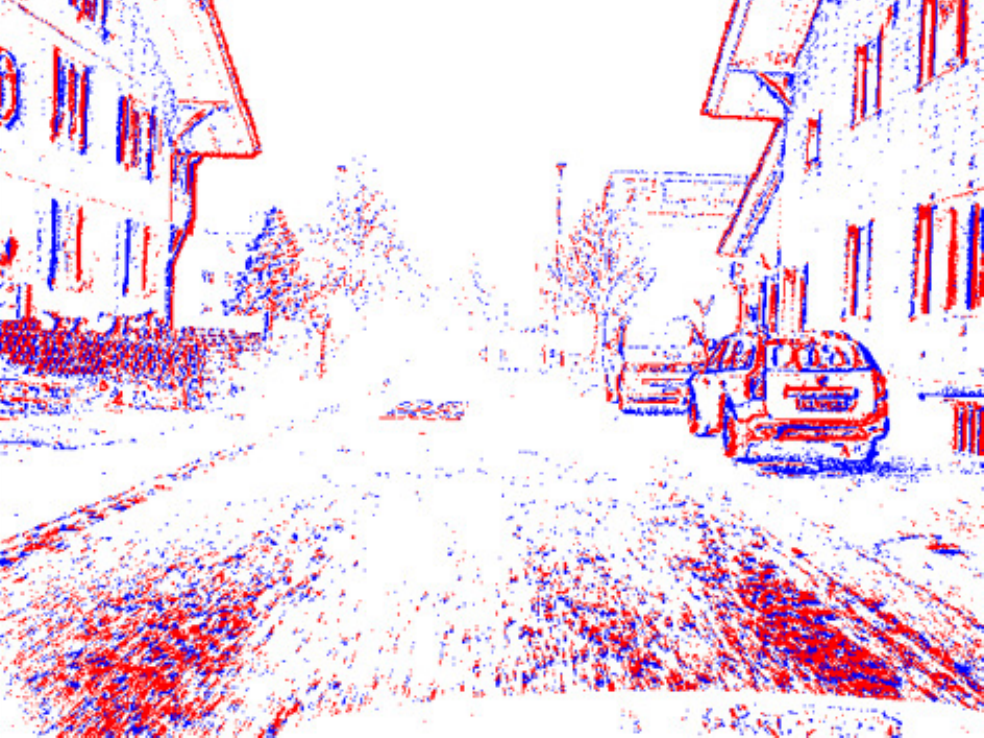}\\
    \end{minipage}
    }
\subfigure[(c) Predicted flow]{
    \begin{minipage}[b]{0.28\linewidth}
        \centering
        \includegraphics[scale=0.166]{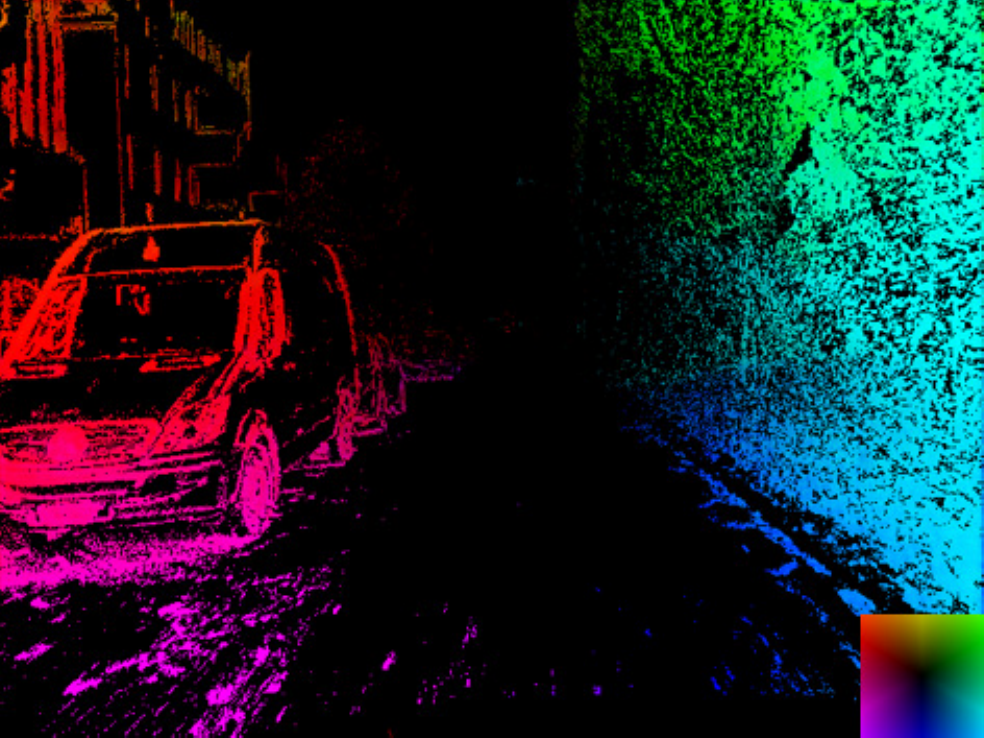}\\
        \vspace{1.6pt}
        \includegraphics[scale=0.166]{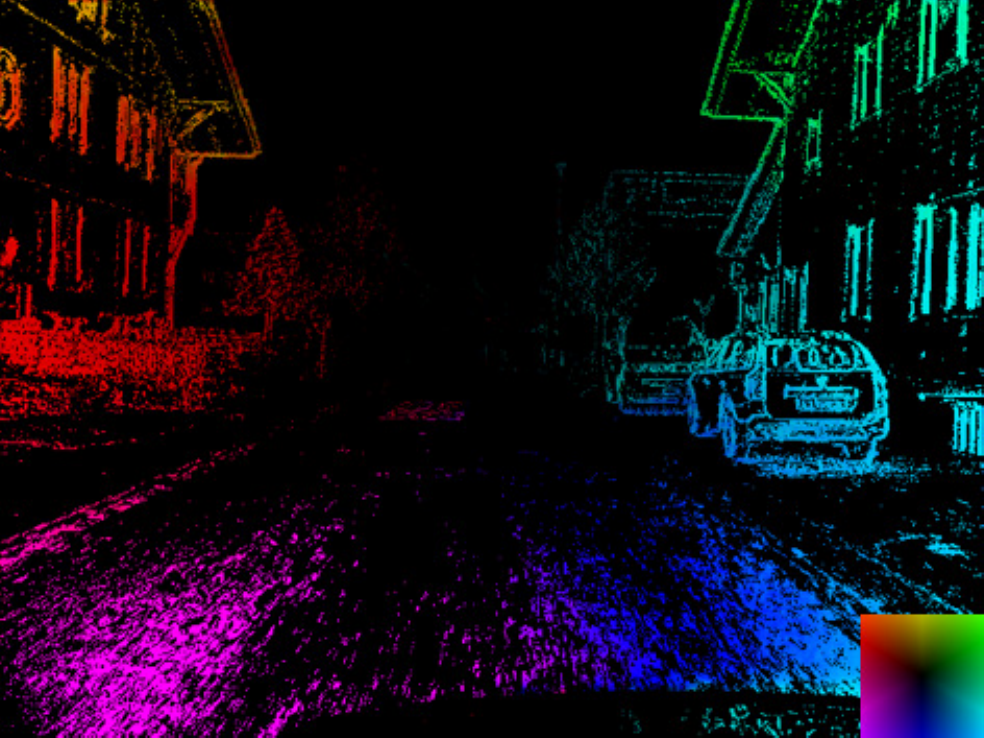}\\
    \end{minipage}
    }
\vspace{-0.17cm}
\caption{\textbf{Qualitative examples on DSEC-Flow sequences.} (a) Image for visualization only. (b) Events for input. (c) Predicted optical flow by our method.}
\label{Fig_intro}
\vspace{-0.8cm}
\end{figure}

Recently, event-based optical flow estimation has gained popularity in the field of robotic vision. Event cameras\cite{gallego2020event}  respond to per-pixel brightness changes independently and output a stream of events asynchronously. Additionally, event cameras offer significant advantages\cite{sanket2020evdodgenet,hou2023fe,forrai2023event} over frame-based cameras, such as high temporal resolution and high dynamic range. Hence, event cameras have emerged as a promising sensor for optical flow estimation.

Most existing learning-based methods for event-based optical flow estimation are inspired by frame-based approaches. For example, early works\cite{zhu2018ev,zhu2019unsupervised} typically follow a matching paradigm between static frames. They convert events into image-like representations and utilize neural networks to learn pixel-level 
correspondences for optical flow estimation. However, their ability to establish temporal correlations among events is limited by the input representation, which fails to fully utilize the spatio-temporal information of events. 
Subsequent works\cite{hagenaars2021self,ding2022spatio,tian2022event,zhuang2024ev} have introduced recurrent neural networks (RNNs) to learn spatio-temporal associations between events triggered by the same object for optical flow estimation. Despite this, these methods struggle to adequately extract event information from both the temporal and spatial dimensions. As the input sequence length increases, they may lead to mismatched spatio-temporal associations, resulting in poor predictions of optical flow over longer time intervals.

In addition, existing unsupervised methods\cite{zhu2019unsupervised,hagenaars2021self,paredes2021back,lee2020spike,lee2022fusion,ding2022spatio,tian2022event,paredes2023taming,zhuang2024ev} for event-based optical flow assume linear motion between events within the loss time window. Only when the input sequence length is short enough or objects move linearly, can adjacent events satisfy the linear motion assumption. Nonetheless, there are numerous complex motions (nonlinear motion\cite{xu2019quadratic,tulyakov2022time}) in real-world cases. Although optical flow produces small errors for each short-time sequence, these errors accumulate as the sequence length increases, leading to degraded performance in longer-time sequences.

To address these challenges, we propose E-NMSTFlow, a novel nonlinear motion-guided and spatio-temporal aware network for unsupervised event-based optical flow estimation. To better learn spatio-temporal data associations, we propose the Spatio-Temporal Motion Feature Aware (STMFA) module. It extracts rich motion information from the temporal and spatial dimensions by aggregating previous features. Next, we present the Adaptive Motion Feature Enhancement (AMFE) module. AMFE adaptively selects weights and enhances features through the self-attention mechanism, which refines motion information and alleviates the phenomenon of local feature similarity. Additionally, we propose a novel unsupervised loss that exploits nonlinear motion between events to reduce optical flow errors in long-time sequences. Finally, we demonstrate the effectiveness of our method on the MVSEC\cite{maqueda2018event} and DSEC-Flow\cite{gehrig2021dsec} datasets, which outperforms the state-of-the-art (SOTA) unsupervised learning method TamingCM by 19.20\% on MVSEC and 13.30\% on DSEC-Flow. The qualitative examples on the DSEC-Flow are shown in Fig. \ref{Fig_intro}.

In summary, the contributions of our work are as follows:
\begin{itemize}
    \item We propose E-NMSTFlow, a novel unsupervised network, which leverages rich spatio-temporal information and nonlinear motion between events to capture motion cues for event-based optical flow estimation.
    \item We propose STMFA to aggregate spatio-temporal motion information and introduce AMFE to refine motion features, enhancing the ability to establish spatio-temporal data associations.
    \item We design a nonlinear motion compensation loss that leverages the accurate nonlinear motion between events, facilitating unsupervised learning in the network.
    \item Extensive experiments on the MVSEC and DSEC-Flow datasets demonstrate that our approach achieves SOTA performance among unsupervised learning methods.
\end{itemize}
\vspace{-0.25cm}
\section{Related Works}
The rise of deep learning has significantly advanced the field of event-based optical flow estimation. We provide a general overview of existing works regarding network architecture and learning paradigms.

\textbf{Regarding Network Architecture.} EV-FlowNet\cite{zhu2018ev} was the first to introduce the U-Net architecture\cite{ronneberger2015u} for event-based optical flow estimation. It was subsequently extended to EV-FlowNet+\cite{zhu2019unsupervised}, which simultaneously estimates optical flow, depth and ego-motion using only events. However, the aforementioned recurrent-free methods predict optical flow sequentially using handcrafted input representations that incorporate temporal information, leading to challenges in extracting spatio-temporal information. Subsequently, due to the remarkable ability of recurrent networks to process sequential inputs, recurrent-based methods have gained popularity. Hagenaars et al.\cite{hagenaars2021self} proposed a recurrent version of EV-FlowNet (ConvGRU-EV-FlowNet) that enhances the network's spatio-temporal modeling capability. Ding et al.\cite{ding2022spatio} proposed STE-FlowNet, which effectively extracts spatio-temporal information by incorporating correlation layers. Later, Tian et al.\cite{tian2022event} introduced ET-FlowNet, which combines RNN with Transformer\cite{vaswani2017attention} to achieve better long-term modeling performance. Recently, EV-MGRFlowNet\cite{zhuang2024ev} estimated optical flow by utilizing previous hidden states and prior flows. However, these methods still do not fully explore the spatio-temporal associated information of events, which is necessary for long-time sequences estimation.

\textbf{Regarding Learning Paradigms.} Learning-based methods can be categorized into supervised and unsupervised methods according to the learning paradigms. Supervised methods use the ground truth (GT) optical flow of events as supervisory signals. However, it is non-trivial to obtain accurate GT optical flow at event-equivalent rates. Due to the difficulties in obtaining GT for event-based optical flow, research efforts have shifted from supervised to unsupervised methods. The losses of unsupervised methods can be divided into photometric consistency loss\cite{zhu2018ev,lee2020spike,lee2022fusion,ding2022spatio} and motion compensation loss\cite{zhu2019unsupervised,mitrokhin2018event,hagenaars2021self,tian2022event,zhuang2024ev}. Photometric consistency loss aims to minimize the photometric error of grayscale images between event streams. However, the requirement for additional images and the motion blur in these images often lead to inaccurate supervisory information.
For motion compensation loss, Zhu et al.\cite{zhu2019unsupervised} first proposed an event-based motion compensation loss in EV-FlowNet+, which utilizes the motion of events to deblur event images. Subsequently, ConvGRU-EV-FlowNet\cite{hagenaars2021self} and ET-FlowNet\cite{tian2022event} also employed the motion compensation loss. To achieve better event alignment, EV-MGRFlowNet\cite{zhuang2024ev} introduced a hybrid motion compensation loss. However, the aforementioned event-based optical flow estimation methods all assume linear motion between consecutive events within the loss time window, which ignores the potential of high temporal resolution events. Additionally, this assumption accumulates optical flow errors in complex scenes, leading to degraded performance in long-time sequences.

\section{Method}
\subsection{Event Representation}
Event cameras output a series of events that respond to changes of intensity asynchronously. An event $\boldsymbol{e}=(x,y,t,p)$ is triggered at pixel location $(x,y)$ and timestamp $t$ if the log intensity change exceeds a set threshold $C$. Polarity $p=\pm1$ denotes the direction of the intensity change. In our work, we utilize the event count image\cite{maqueda2018event,hagenaars2021self} as the event representation. Specifically, we first divide the event stream into a series of event volumes $\boldsymbol{E}_t=\{\boldsymbol{e}_i\}_{i=1}^{N}\in\mathbb{R}^{N\times 4}$ at fixed intervals. Then, the event volumes are processed into an image-like event representation $\boldsymbol{I}_{t}^{\pm}\in\mathbb{R}^{2\times H\times W}$ as the input of our network, as follows:
\begin{equation}
    \boldsymbol{I}_t^{\pm}(x,y)=\sum_{\boldsymbol{e}_i\in \boldsymbol{E}_t^{\pm}}\left|p_{i}\right|\delta(x-x_i,y-y_i)
\end{equation}
\noindent where $\delta(\cdot)$ denotes the Dirac pulse.
\begin{figure*}[t]
\centering
\includegraphics[trim=5 8 15 10, clip, scale=0.51]{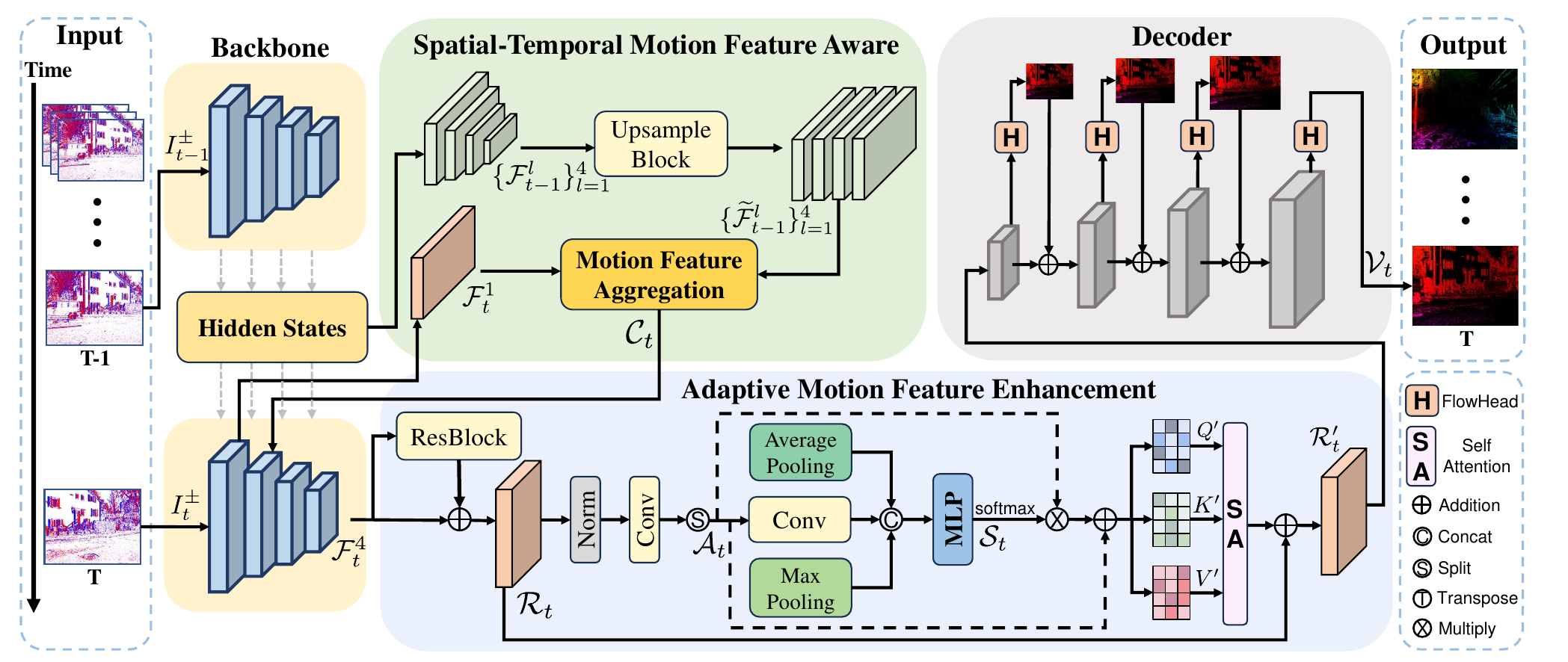}
\caption{\textbf{The overview of our proposed E-NMSTFlow.} First, we extract multi-scale features with a backbone network. Next, Spatio-Temporal Motion Feature Aware (STMFA) module utilizes previous multi-scale features with current low-level features to aggregate spatio-temporal motion information. Then, Adaptive Motion Feature Enhancement (AMFE) module adaptively enhances motion features. Finally, we upsample enhanced motion features and output predicted optical flow. }
\label{framework}
\vspace{-0.5cm}
\end{figure*}
\vspace{-0.2cm}
\subsection{Framework Overview}
Fig. \ref{framework} illustrates our proposed method for event-based optical flow estimation, which primarily consists of four parts: (1) a backbone network for feature extraction; (2) the Spatio-Temporal Motion Feature Aware (STMFA) module; (3) the Adaptive Motion Feature Enhancement (AMFE) module; and (4) a decoder module for optical flow prediction.

Overall, our E-NMSTFlow takes the event count image $\boldsymbol{I}_{t}^{\pm}\in\mathbb{R}^{2\times H\times W}$ as input and outputs optical flow $\mathcal{V}_t=\{v_x,v_y\}\in\mathbb{R}^{2\times H\times W}$. First, we employ four recurrent convolutional layers followed by ConvGRU, which extract multi-scale features $\{{\mathcal{F}_t^l}\}_{l=1}^4\in\mathbb{R}^{C_1\times2^{l-1}\times\frac H{2^l}\times\frac W{2^l}}$ from event count images $\boldsymbol{I}_{t}^{\pm}$, where $C_1$ is the channel dimension of the first scale features and we set it to 64. Next, we introduce the STMFA module, which utilizes multi-scale features to explore spatio-temporal motion information. Subsequently, we adaptively enhance the motion features using the AMFE module. Finally, we utilize a decoder module to upsample the enhanced motion features $\mathcal{R}_t^{\prime}\in\mathbb{R}^{8\times C_1\times\frac H{16}\times\frac W{16}}$ and predict nonlinear motion parameters $\boldsymbol{\theta}\in\mathbb{R}^{3\times H\times W}$, which are used for the nonlinear motion compensation loss computation. Then, the optical flow $\mathcal{V}_t\in\mathbb{R}^{2\times H\times W}$ is further calculated by predicted nonlinear motion parameters. 
\subsection{Spatio-Temporal Motion Feature Aware}
Our backbone network for feature extraction is a multi-scale recurrent neural network based on ConvGRU\cite{ballas2016delving}, which is widely adopted in event-based optical flow estimation\cite{hagenaars2021self,tian2022event,zhuang2024ev}. However, we observe that previous methods \cite{hagenaars2021self,tian2022event} based on the backbone network ignore data associations at different scales in the temporal dimension. Thus, low-level features at the current moment may significantly lose motion information from previous higher-level features, which contain fine motion cues related to pixel-level correspondences.

\begin{figure}[t]
\centering
\includegraphics[scale=0.5]
{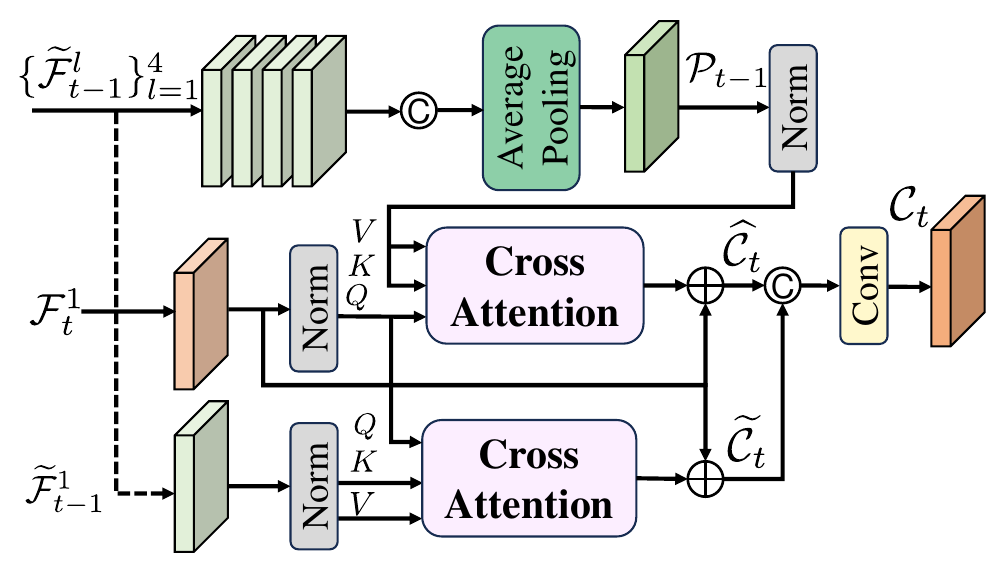}
\caption{\textbf{Details of Motion Feature Aggregation.}}
\vspace{-0.5cm}
\label{figure3_detail}
\end{figure}

To make full use of the spatio-temporal information of events and achieve more robust optical flow estimation for long-time sequences, we propose the Spatio-Temporal Motion Feature Aware (STMFA) module, as shown in Fig. \ref{framework}. Specifically, we first upsample multi-scale motion features $\{{\mathcal{F}_{t-1}^l}\}_{l=1}^4$ at different scales from the previous hidden states to get $\{{\mathcal{\widetilde{F}}_{t-1}^l}\}_{l=1}^4$. Then, we propose the Motion Feature Aggregation module to aggregate multiple motion features, and the details are shown in Fig. \ref{figure3_detail}. To preserve both the global information and local details of motion features from the previous time step, we concatenate features in the channel dimension and fuse previous motion features through Average Pooling to obtain fused features $\mathcal{P}_{t-1}$.
\begin{equation}
    \begin{aligned}
        \mathcal{P}_{t-1} = \mathrm{AvgPool}(\mathrm{Concat}(\{\mathcal{\widetilde{F}}_{t-1}^l\}_{l=1}^4))
    \end{aligned}
\end{equation}
Next, we introduce Cross-Attention\cite{chen2021crossvit} to aggregate multiple features at different moments. Specifically, to fully retain the global motion information, we aggregate the previous features at the first scale $\mathcal{\widetilde{F}}_{t-1}^1$ and previous motion features $\mathcal{P}_{t-1}$ with current features at the first scale $\mathcal{F}_t^1$, respectively. The above process can be expressed as:
\begin{equation}
    \begin{aligned}
        \mathcal{\widehat{C}}_{t} = \mathrm{CA}(\mathrm{LN}(\mathcal{F}_t^1), \mathrm{LN}(\mathcal{P}_{t-1}), \mathrm{LN}(\mathcal{P}_{t-1}))+\mathcal{F}_t^1
    \end{aligned}
\end{equation}
\begin{equation}
    \begin{aligned}
        \mathcal{\widetilde{C}}_{t} = \mathrm{CA}(\mathrm{LN}(\mathcal{F}_t^1), \mathrm{LN}(\mathcal{\widetilde{F}}_{t-1}^1), \mathrm{LN}(\mathcal{\widetilde{F}}_{t-1}^1))+\mathcal{F}_t^1
    \end{aligned}
\end{equation}
where $\mathrm{CA}$ represents Cross-Attention. LN means a normalization layer\cite{ba2016layer}. 
We concatenate the aggregated features $\mathcal{\widehat{C}}_{t}$ and $\mathcal{\widetilde{C}}_{t}$, and apply a convolution operation to obtain the final feature $\mathcal{C}_{t}$, which is used for feature extraction at the next scale, as follows:
\begin{equation}
    \begin{aligned}
        \mathcal{C}_{t} = \mathrm{Conv}(\mathrm{Concat}(\mathcal{\widehat{C}}_{t}, \mathcal{\widetilde{C}}_{t}))
    \end{aligned}
\end{equation}
The aggregated features contain accurate global motion features and fine-grained local feature information. Therefore, our STMFA can utilize rich spatio-temporal motion features to establish data associations, achieving more robust optical flow estimation in long-time sequences.
\subsection{Adaptive Motion Feature Enhancement}
We observe that local motion feature similarity occurs under high-speed motion, which affects the learning and convergence of the network. Inspired by this observation, an Adaptive Motion Feature Enhancement (AMFE) module is proposed to focus on meaningful motion features, aiming to alleviate the local feature similarity phenomenon. Specifically, as shown in Fig. \ref{framework}, the feature map $\mathcal{F}_t^4$ extracted through the backbone network is input into two residual blocks for further feature extraction, and the residual features $\mathcal{R}_t$ are obtained. Then, we use a normalization layer\cite{ba2016layer} and a 2D convolutional layer to generate three attention vectors $\mathcal{A}_t\in\{Q, K, V\}$. To enhance motion features in attention vectors, we adaptively select meaningful attention vectors. This process can be expressed by the following formula:
\begin{equation}
    \begin{aligned}
        \mathcal{S}_t=\mathrm{MLP}(\mathrm{Concat}(\mathrm{Conv}(\mathcal{A}_t),\mathrm{Max}(\mathcal{A}_t),\mathrm{Avg}(\mathcal{A}_t)))
    \end{aligned}
\end{equation}
\begin{equation}
    \begin{aligned}
        \mathcal{A}_t^{\prime} = \mathrm{Softmax}(\mathcal{S}_t)\mathcal{A}_t+\mathcal{A}_t
    \end{aligned}
\end{equation}
where Max and Avg represent Max Pooling and Average Pooling respectively, and $\mathcal{A}_t^\prime\in\{Q^{\prime}, K^{\prime}, V^{\prime}\}$. Finally, we calculate enhanced motion features $\mathcal{R}_t^{\prime}$ by the selected attention vectors ${Q}^{\prime}$, ${K}^{\prime}$, ${V}^{\prime}$ as follows:
\begin{equation}
\label{Attention weight}
    \begin{aligned}
        \mathcal{R}_t^{\prime} = \mathrm{SA}({Q}^{\prime},{K}^{\prime},{V}^{\prime})+\mathcal{R}_t
    \end{aligned}
\end{equation}
\vspace{-0.05cm}
where $\mathrm{SA}$ represents Self-Attention.
\vspace{-0.05cm}
\subsection{Nonlinear Motion Compensation Loss}
Inspired by \cite{shiba2022event}, we argue that the motion between consecutive events is nonlinear. Based on this, we propose a nonlinear motion compensation loss to make full use of the nonlinear characteristics of events and achieve better unsupervised learning of the network, thereby improving the performance of our method in complex scenes. According to \cite{hagenaars2021self}, accurate optical flow information is encoded in the spatio-temporal misalignment among the events triggered by the same portion of a moving edge. To retrieve it, we employ the nonlinear motion model to compensate for this motion by warping events. Specifically, we use the nonlinear motion model $\mathcal{W}$to obtain the warped event volume $\boldsymbol{E}_t^{\prime}=\{\boldsymbol{e}_i^{\prime}\}_{i=1}^{N}$ at the reference time $t_r$, as follow:
\begin{equation}
    \begin{aligned}
    \boldsymbol{e}_i\to \boldsymbol{e}_i^\prime:\quad\mathbf{x}_i'=\mathcal{W}\left(\mathbf{x}_i, t_i; \boldsymbol{\theta}\right)
    \end{aligned}
\end{equation}
where $\boldsymbol{\theta}=(\mathbf{v}, \omega)$ is estimated by our network. $\mathbf{x}=(x, y)$ means the pixel coordinate of an event. The nonlinear motion model $\mathcal{W}$ is parameterized by a translation parameter $\mathbf{v}=(v_x, v_y)$ and an in-plane rotation parameter $\omega$:
\begin{equation}
            \begin{aligned}
		\begin{pmatrix}
			\mathbf{x}_i^{\prime}\\1
		\end{pmatrix}
		\sim
		\begin{pmatrix}
			{\mathrm{R}}(t_i{\omega}) & t_i\mathbf{v}\\\mathbf{0}^\top & 1
		\end{pmatrix}
		\begin{pmatrix}
			\mathbf{x}_i\\1
		\end{pmatrix}
            \end{aligned}
	\end{equation}
 where $\mathrm{R}$ is parameterized by using exponential coordinates (Rodrigues rotation formula\cite{Murray1994AMI,Gallego_2014}). To obtain the final optical flow by predicted nonlinear motion parameters $\boldsymbol{\theta}$. The optical flow is calculated by Eq. \ref{flow}.
 	\begin{equation}
		\begin{aligned}\label{flow}
			\mathcal{V}_i&=\frac{\partial \mathbf{x}_i^{\prime}}{\partial t}
   =\mathrm{R}(\frac{\pi}{2}+t{\omega})\mathbf{x}_i\omega+\mathbf{v}
		\end{aligned}
	\end{equation}
 Then, we need to formulate a loss function to measure how well our nonlinear motion model compensates for the motion of events. Following \cite{zhuang2024ev}, we use the average timestamp of the image of warped events (IWE) $T_{p^{\prime}}$ and the exponential count IWE $E_{p^ {\prime}}$ to measure compensation quality. The lower this metric, the better the model compensates for the motion of events and the more accurate the estimated optical flow is. $T_{p^{\prime}}$ and $E_{p^ {\prime}}$ are represented as follows:
\begin{equation}
		\begin{aligned}
			T_{p^{\prime}}(\mathbf{x};\boldsymbol{\theta}|t_r)=\frac{\sum_j\kappa(x-x_j^{\prime})\kappa(y-y_j^{\prime})t_j}{\sum_j\kappa(x-x_j^{\prime})\kappa(y-y_j^{\prime})+\epsilon}
		\end{aligned}
	\end{equation}
  	\begin{equation}
		\begin{aligned}
		 E_{p^{\prime}}(\mathbf{x};\boldsymbol{\theta}|t_r)=\exp(-\alpha\sum_j\kappa(x-x_j^{\prime})\kappa(y-y_j^{\prime}))
		\end{aligned}
	\end{equation}
 where $\kappa$ is the linear sampling kernel, $\kappa(a)=\max(0,1-|a|)$, $j\in\{i|p_i=p'\}$, and $\alpha$ is the saturation factor.
Thus, we can derive the loss functions with $T_{p^{\prime}}$ and $E_{p^ {\prime}}$ as follows:
\begin{equation}
		\begin{aligned}
		 \mathcal{L}_{\mathrm{AT}}(t_r)=\frac{\sum_{\mathbf{x}}T_+(\mathbf{x};\boldsymbol{\theta}|t_r)^2+T_-(\mathbf{x};\boldsymbol{\theta}|t_r)^2}{\sum_{\mathbf{x}}[n(\mathbf{x}^{\prime})>0]+\epsilon}
		\end{aligned}
	\end{equation}
\begin{equation}
		\begin{aligned}
        \mathcal{L}_{\mathrm{EC}}(t_r)=\frac{N_P}{\sum_{\mathbf{x}}E_+(\mathbf{x};\boldsymbol{\theta}|t_r)}
        + \frac{N_P}
        {\sum_{\mathbf{x}}E_-(\mathbf{x};\boldsymbol{\theta}|t_r}
		\end{aligned}
\end{equation}
where $n(\mathbf{x}^{\prime})$ denotes per-pixel event counts of IWE, and $N=H\times W$ is the total number of pixels. Besides, we warp events both in forward and backward fashions to avoid temporal scaling issues during backpropagation\cite{hagenaars2021self}, as follows:
\begin{equation}
		\begin{aligned}
    \mathcal{L}_{\mathrm{AT}}=\mathcal{L}_{\mathrm{AT}}(t_r^{fw})
  +\mathcal{L}_{\mathrm{AT}}(t_r^{bw})
		\end{aligned}
\end{equation}
\begin{equation}
		\begin{aligned}
    \mathcal{L}_{\mathrm{EC}}=\mathcal{L}_{\mathrm{EC}}(t_r^{fw})
  +\mathcal{L}_{\mathrm{EC}}(t_r^{bw})
		\end{aligned}
\end{equation}
\begin{equation}
	\begin{aligned}
  \mathcal{L}_{\mathrm{NLMC}}=\mathcal{L}_{\mathrm{AT}}
  + \lambda_1\mathcal{L}_{\mathrm{EC}} + \lambda_2\mathcal{L}_{\mathrm{smooth}}
	\end{aligned}
\end{equation}
where $\lambda_1$ and $\lambda_2$ balance the different loss terms.
\section{Experiments}
\subsection{Experimental Setup}
\textbf{Datasets.} Following previous works\cite{paredes2023taming,wu2024lightweight,gehrig2021raft}, we conduct experiments on two event-based datasets MVSEC\cite{maqueda2018event} and DSEC-Flow\cite{gehrig2021dsec}. For MVSEC, considering the generalization of our network for complex scenes, we follow works\cite{hagenaars2021self,tian2022event,zhuang2024ev} and train our network on the UZH-FPV drone racing dataset\cite{delmerico2019we}. Then, we test it on four different sequences from the MVSEC dataset with ground truth corresponding to a time interval of dt=1 and dt=4 gray images. 
For DSEC-Flow, we train our model on the official training set and evaluate it on the DSEC-Flow benchmark.
\begin{table*}[t]

\centering
\caption{\textbf{Comparison of our method against state-of-the-art unsupervised methods on MVSEC.}}
\label{TableMVSEC}
\resizebox{\linewidth}{!}{
\begin{tabular}{c l c c c c c c c c}
\toprule[0.05cm]
~ & \multirow{2}{*}{Methods} & \multicolumn{2}{c}{indoor\_flying1} & \multicolumn{2}{c}{indoor\_flying2} & \multicolumn{2}{c}{indoor\_flying3} & \multicolumn{2}{c}{outdoor\_day1} \\ 
\cline{3-10}\rule{0pt}{8pt}
~ & ~ & AEE↓ & \%Out↓ & AEE↓ & \%Out↓ & AEE↓ & \%Out↓ & AEE↓ & \%Out↓ \\ 
\hline\rule{0pt}{8pt}
~ & EV-Flownet+\cite{zhu2019unsupervised} & 0.58 & \textbf{0.00} & 1.02 & 4.00 & 0.87 & 3.00 & 0.32 & \textbf{0.00}  \\
~ & STE-FlowNet\cite{ding2022spatio} & 0.57 & \underline{0.10} & 0.79 & \textbf{1.60} & 0.72 & 1.30 & 0.42 & \textbf{0.00} \\
\multirow{2}{*}{dt=1} & ET-FlowNet\cite{tian2022event}  & 0.57 & 0.53 & 1.20 & 8.48 & 0.95 & 5.73 & 0.39 & 0.12 \\  
~ & ConvGRU-EV-FlowNet\cite{hagenaars2021self} & 0.60 & 0.51 & 1.17 & 8.06 & 0.93 & 5.64 & 0.47 & 0.25 \\ 
~ & EV-MGRFlowNet\cite{zhuang2024ev} & \underline{0.41} & 0.17 & \underline{0.70} & 2.35 & \underline{0.59} & \underline{1.29} & \underline{0.28} & \underline{0.02}\\
~ & TamingCM\cite{paredes2023taming} & 0.44 & \textbf{0.00} & 0.88 & 4.51 & 0.70 & 2.41 & \textbf{0.27} & 0.05 \\
~ & Ours  & \textbf{0.39} & 0.13 & \textbf{0.67} & \underline{2.15} & \textbf{0.55} & \textbf{1.14} & \textbf{0.27} & \underline{0.02}\\ 
\hline
\hline\rule{0pt}{8pt}
~ & EV-Flownet+\cite{zhu2019unsupervised} & 2.18 & 24.20 & 3.85 & 46.80 & 3.18 & 47.80 & 1.30 & 9.70  \\ 
~ & STE-FlowNet\cite{ding2022spatio} & 1.77 & 14.70  & 2.52 & 26.10 & 2.23 &22.10 & \textbf{0.99} & \textbf{3.90} \\
\multirow{2}{*}{dt=4} & ET-FlowNet\cite{tian2022event}   & 2.08 & 20.02 & 3.99 & 41.33 & 3.13 & 31.70 & 1.47 & 9.17 \\
~ & ConvGRU-EV-FlowNet\cite{hagenaars2021self} & 2.16 & 21.51 & 3.90 & 40.72 & 3.00 & 29.60 & 1.69 & 12.50 \\ 
~ & EV-MGRFlowNet\cite{zhuang2024ev} & \underline{1.50} & \underline{8.67} & \underline{2.39} & \underline{23.70} & \underline{2.06} & \underline{18.00} & 1.10 & 6.22\\ 
~ & TamingCM\cite{paredes2023taming} & 1.71 & 12.92 & 3.05 & 31.80 & 2.46 & 24.69 & 1.16 & 5.87 \\
~ & Ours  & \textbf{1.42} & \textbf{6.93} & \textbf{2.26} & \textbf{20.09} & \textbf{1.85} & \textbf{13.68}& \underline{1.04} & \underline{4.50} \\ 
\toprule[0.05cm]
\end{tabular}
}
\label{table_MAP}
\vspace{-0.5cm}
\end{table*}
\begin{figure}[t]
\centering
\subfigure[]{
\hspace{-8pt}
\rotatebox{90}{~~~~~~~~~~~~~~~~~~~~~~~~~~~~~~~~~~\footnotesize ConvGRU-}
\rotatebox{90}{~~~~~\footnotesize \text{Ours} ~~~~~~~~~~\footnotesize TamingCM ~~~~\footnotesize EV-FlowNet ~~~~~~~~~\footnotesize GT ~~~~~~~~~~~~~~\footnotesize Events}
\begin{minipage}[b]{.182\linewidth}
\centering
\includegraphics[scale=0.135]{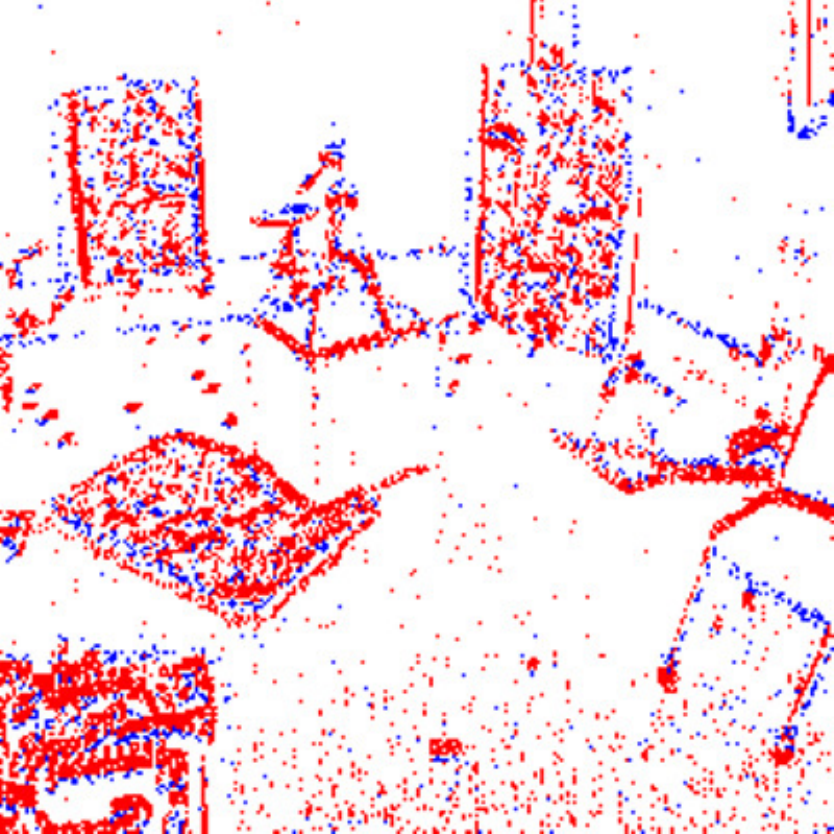}\\
\vspace{1pt}
\includegraphics[scale=0.135]{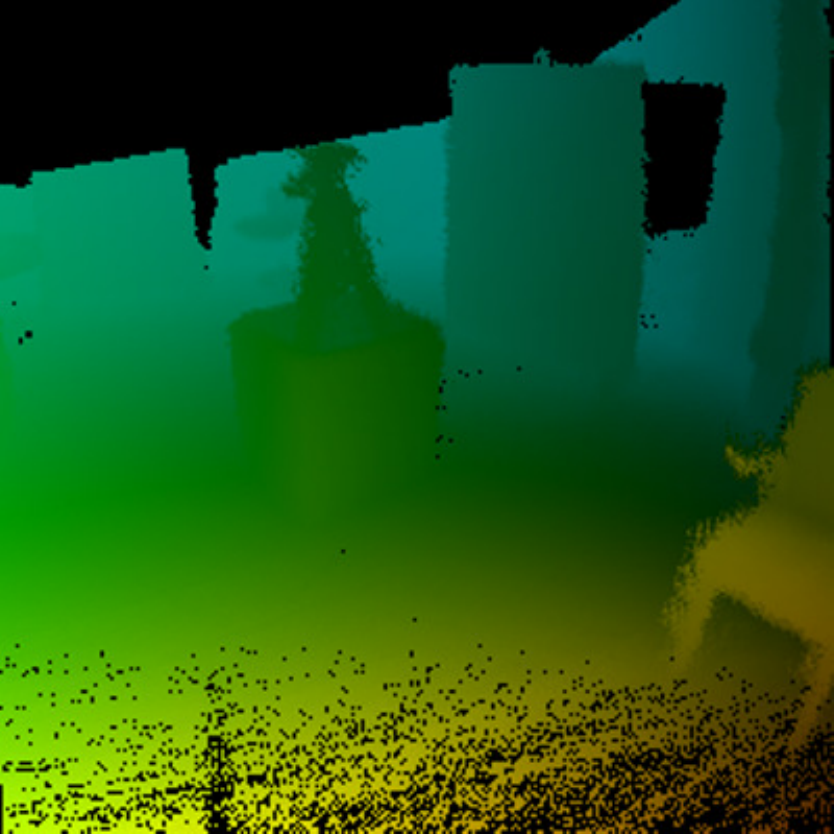}\\
 \vspace{1pt}
\includegraphics[scale=0.135]{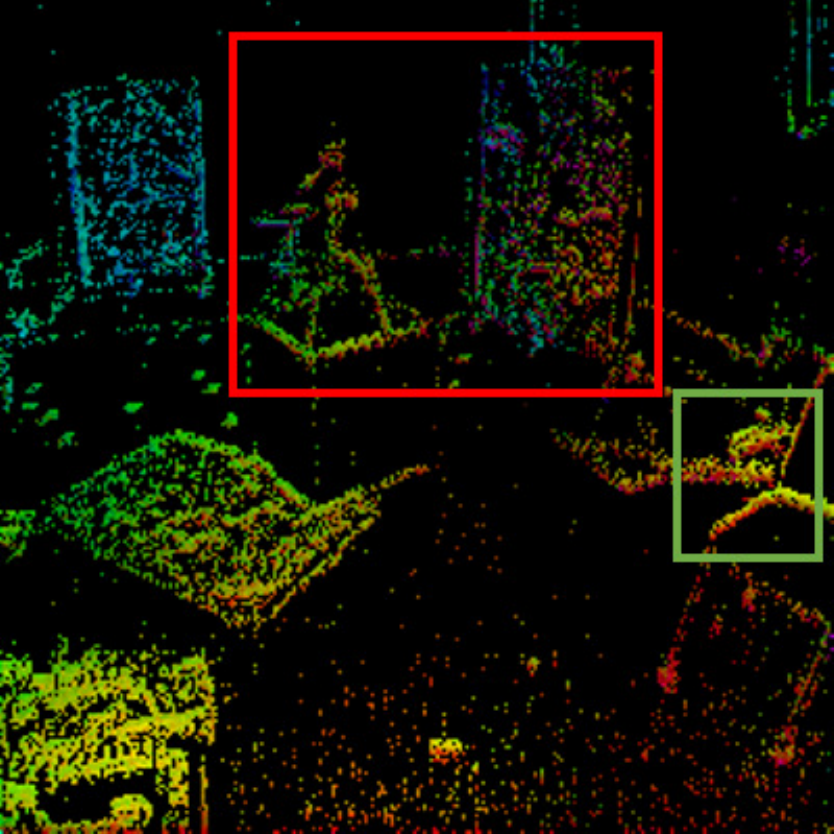}\\
 \vspace{1pt}
\includegraphics[scale=0.135]{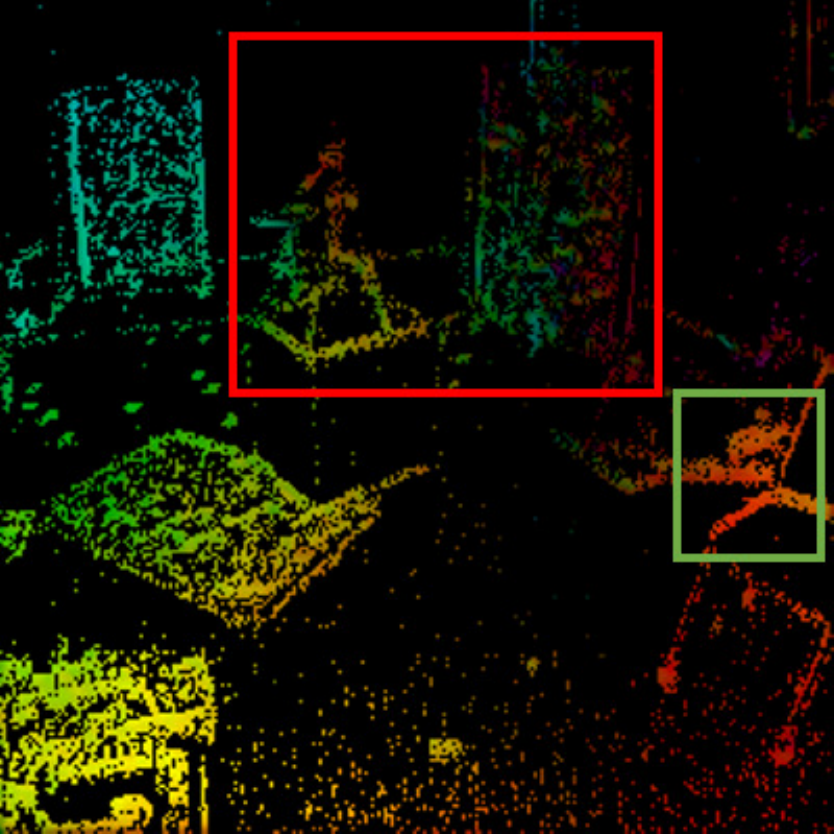}\\
 \vspace{1pt}
\includegraphics[scale=0.135]{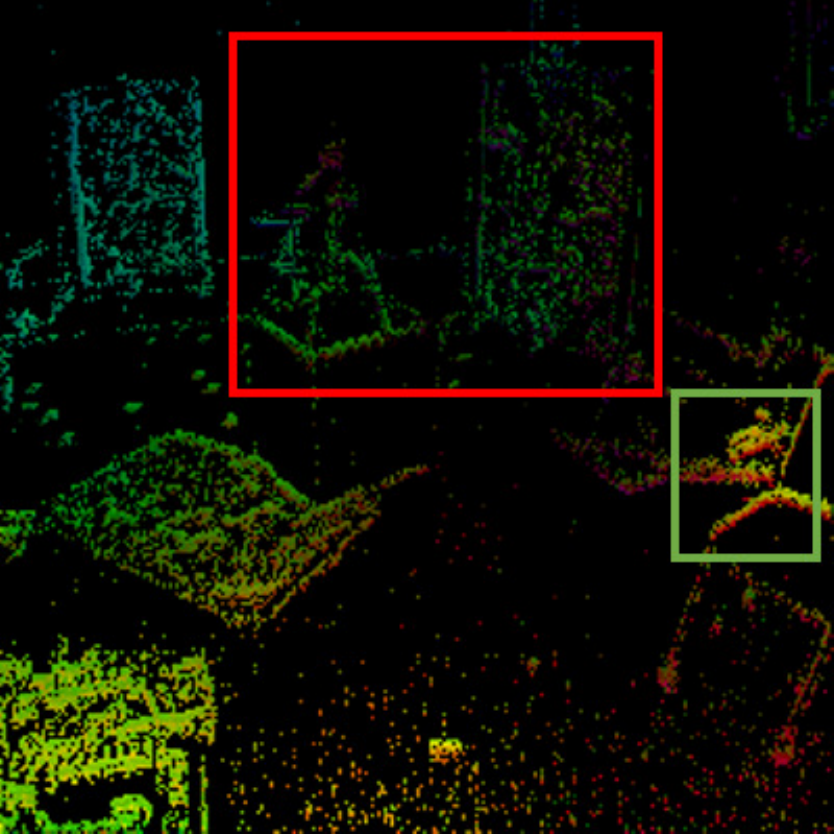}\\
\end{minipage}
}
\subfigure[]{
\begin{minipage}[b]{.182\linewidth}
\centering
\includegraphics[scale=0.135]{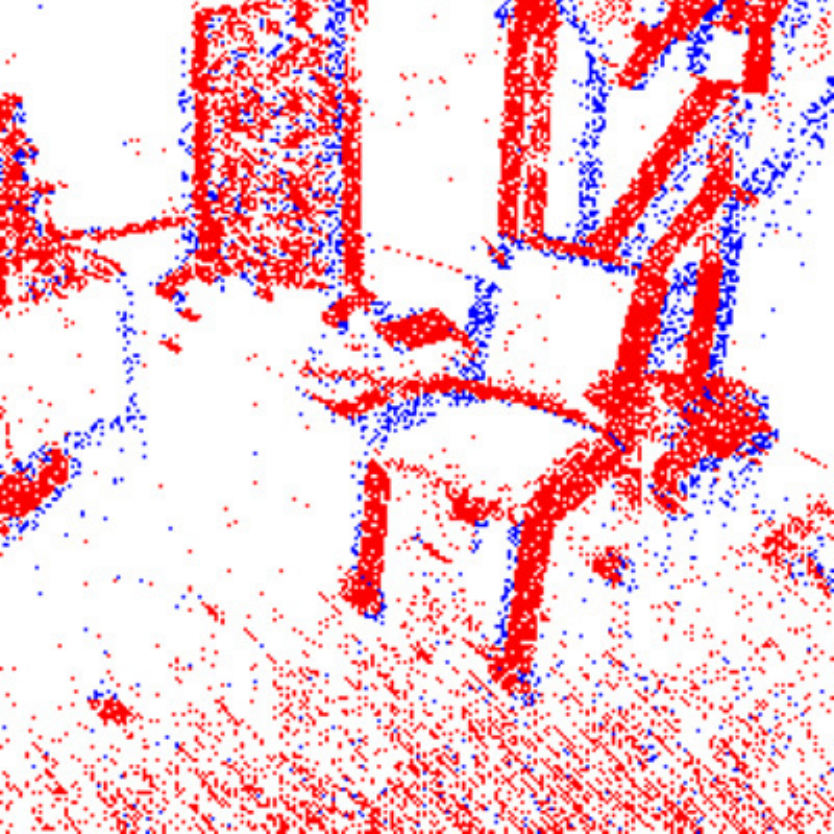}\\
\vspace{1pt}
\includegraphics[scale=0.135]{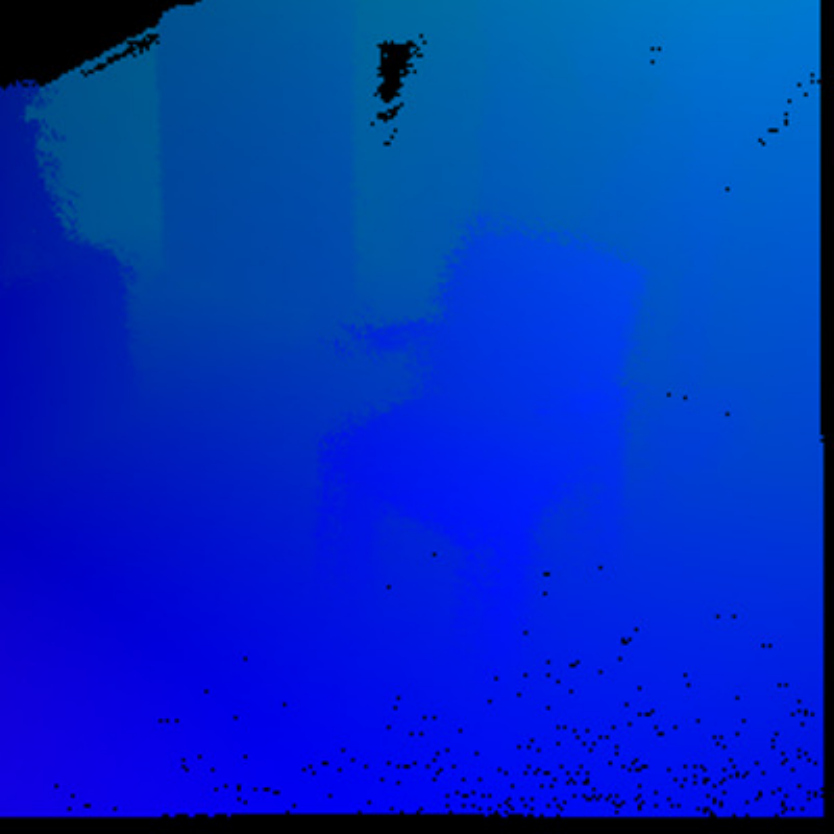}\\
\vspace{1pt}
\includegraphics[scale=0.135]{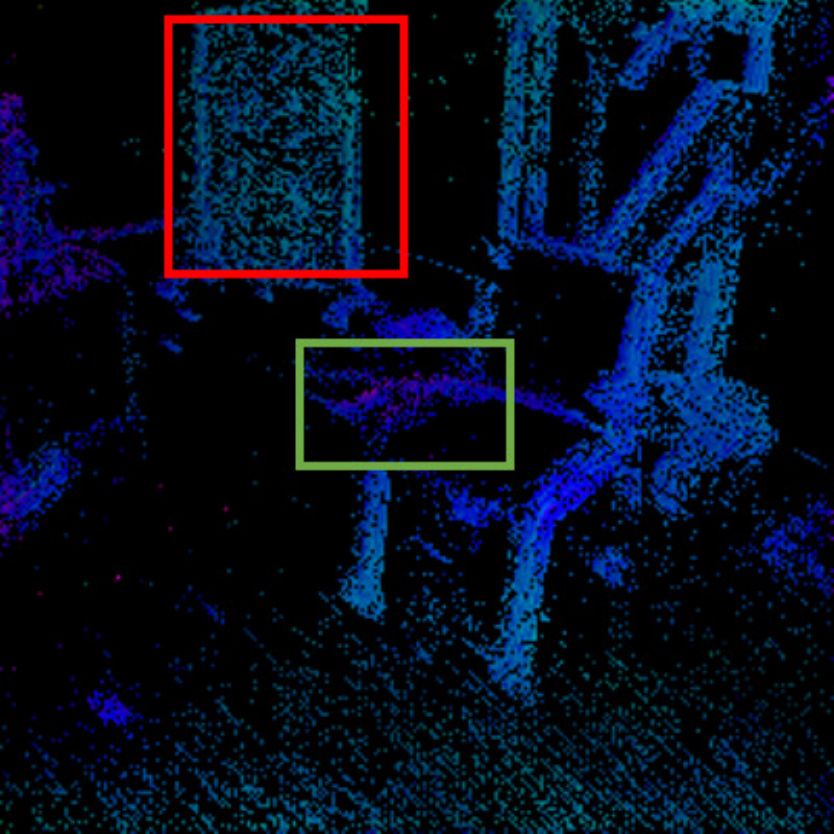}\\
\vspace{1pt}
\includegraphics[scale=0.135]{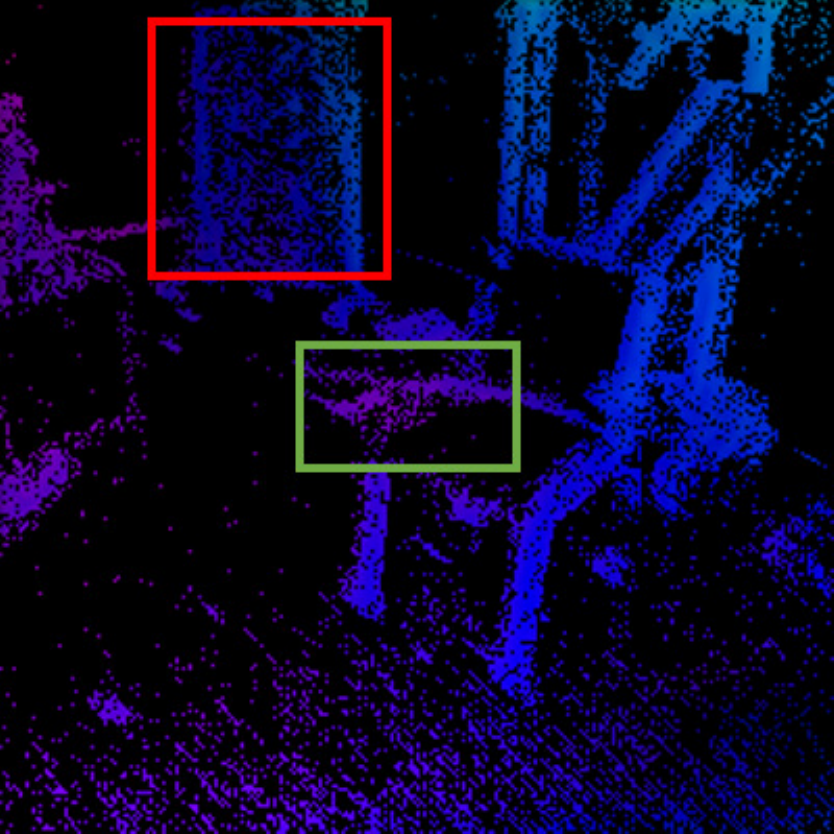}\\
\vspace{1pt}
\includegraphics[scale=0.135]{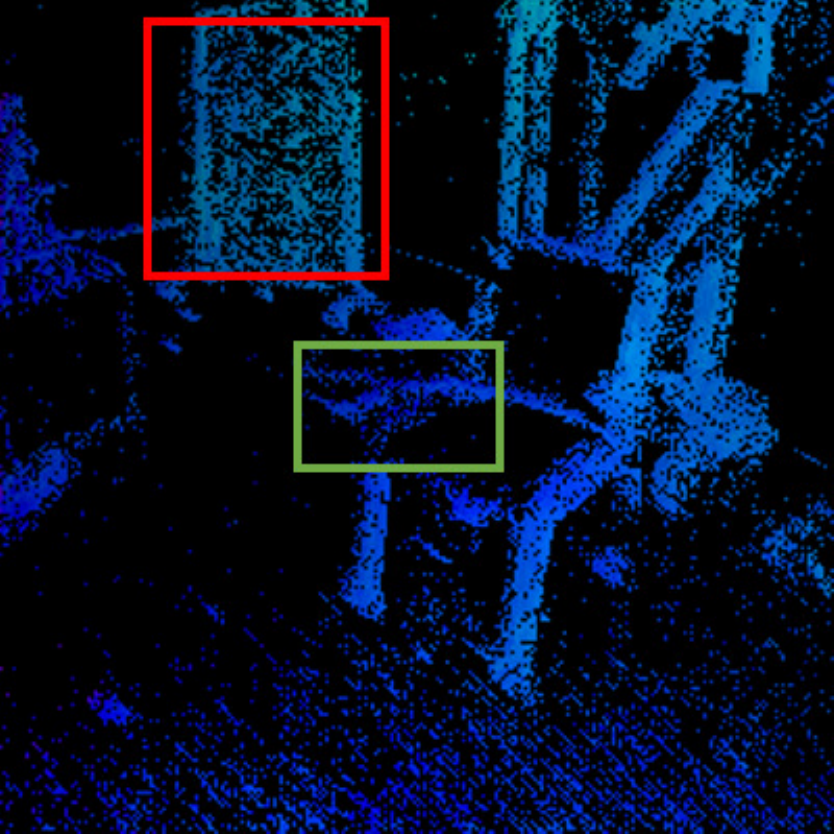}\\
\end{minipage}
}
\subfigure[]{
    \begin{minipage}[b]{.182\linewidth}
        \centering
        \includegraphics[scale=0.135]{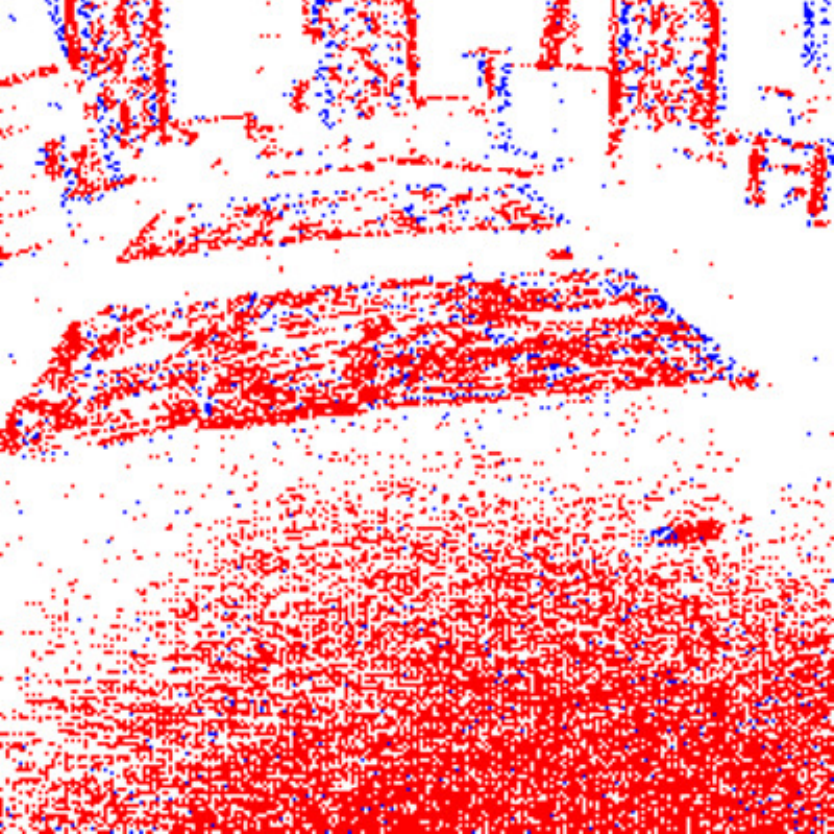}\\
        \vspace{1pt}
        \includegraphics[scale=0.135]{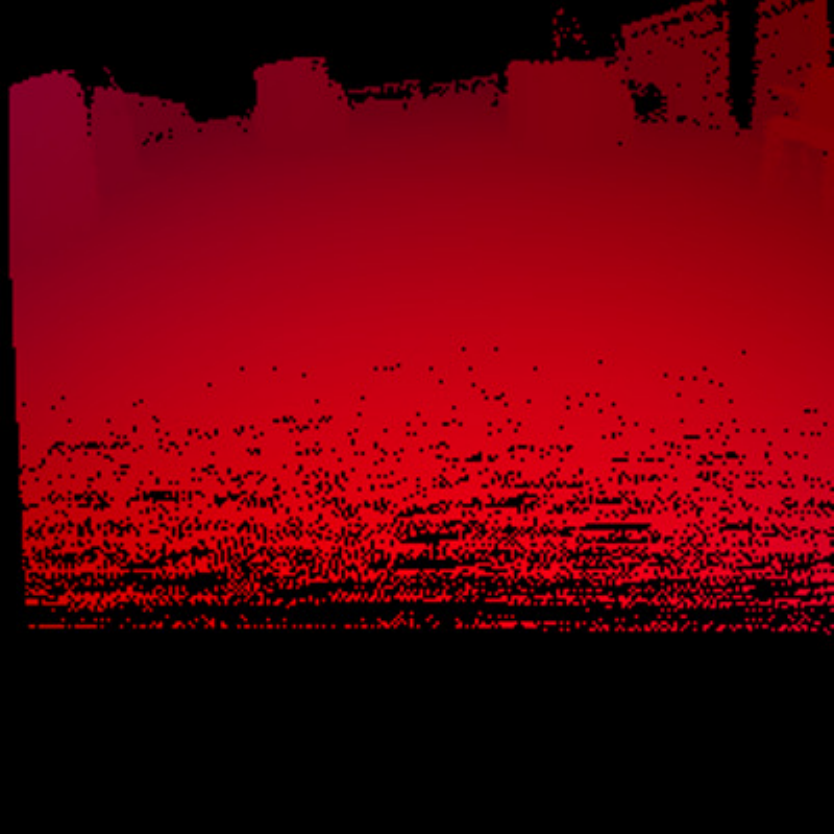}\\
        \vspace{1pt}
        \includegraphics[scale=0.135]{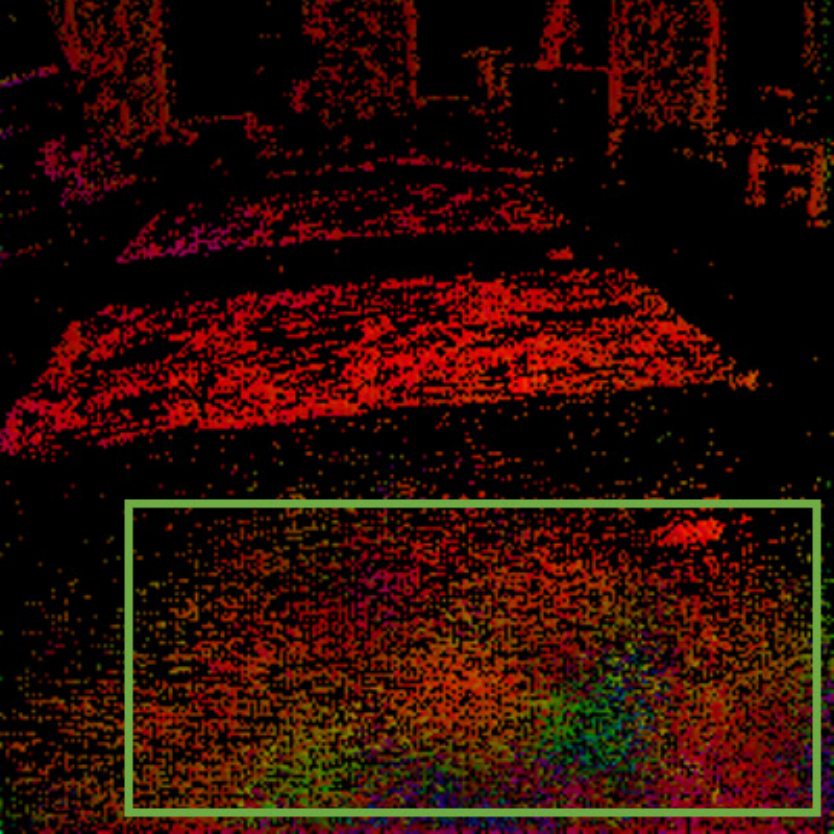}\\
        \vspace{1pt}
        \includegraphics[scale=0.135]{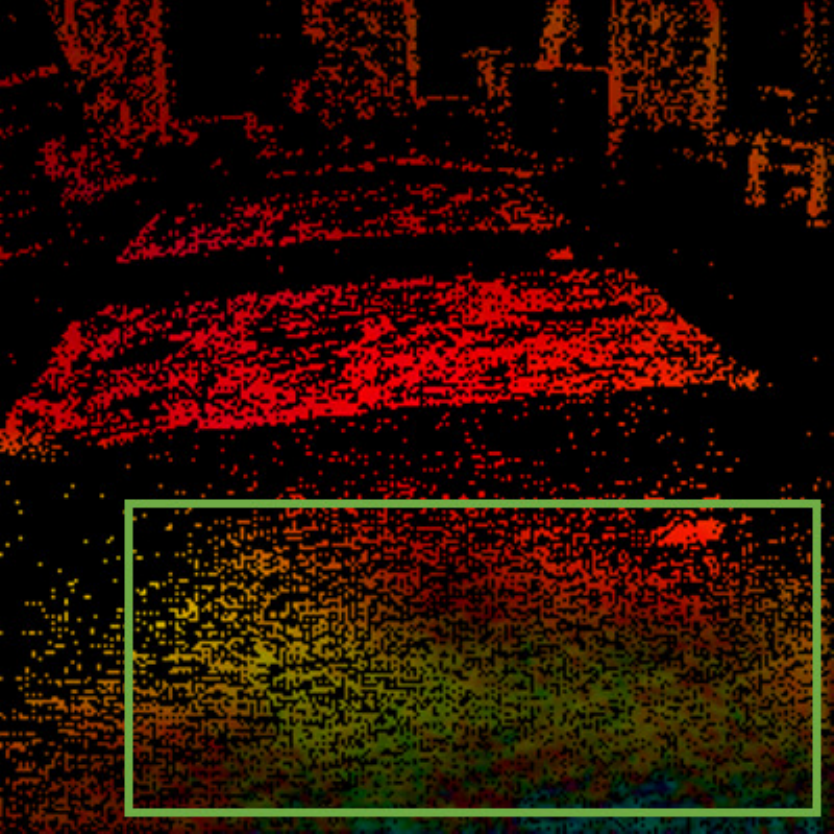}\\
        \vspace{1pt}
        \includegraphics[scale=0.135]{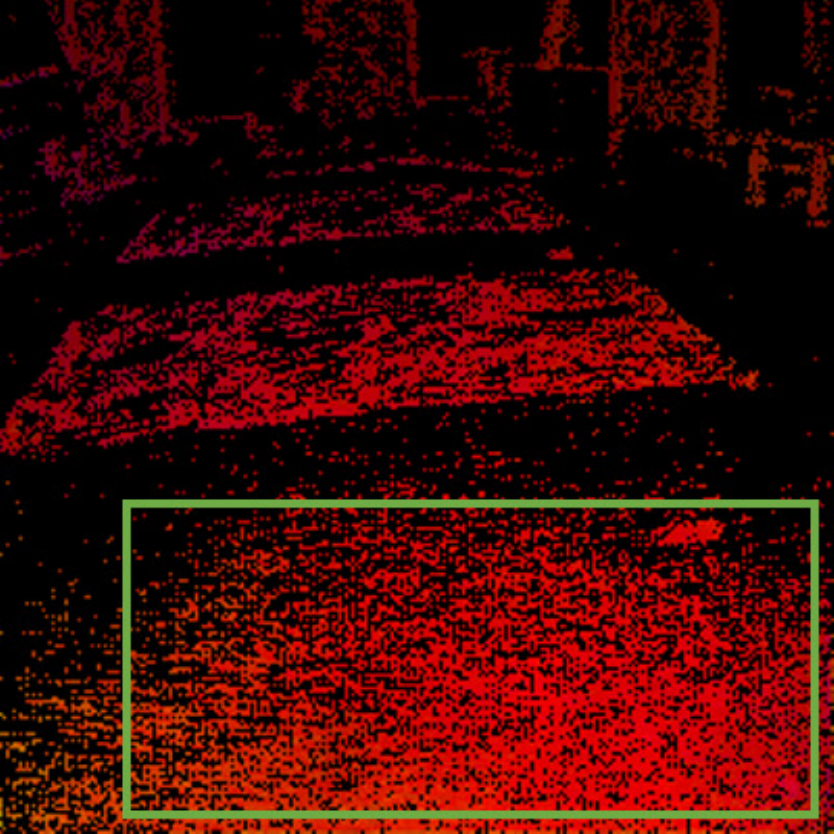}\\
    \end{minipage}
}
\subfigure[]{
    \begin{minipage}[b]{.182\linewidth}
        \centering
        \includegraphics[scale=0.135]{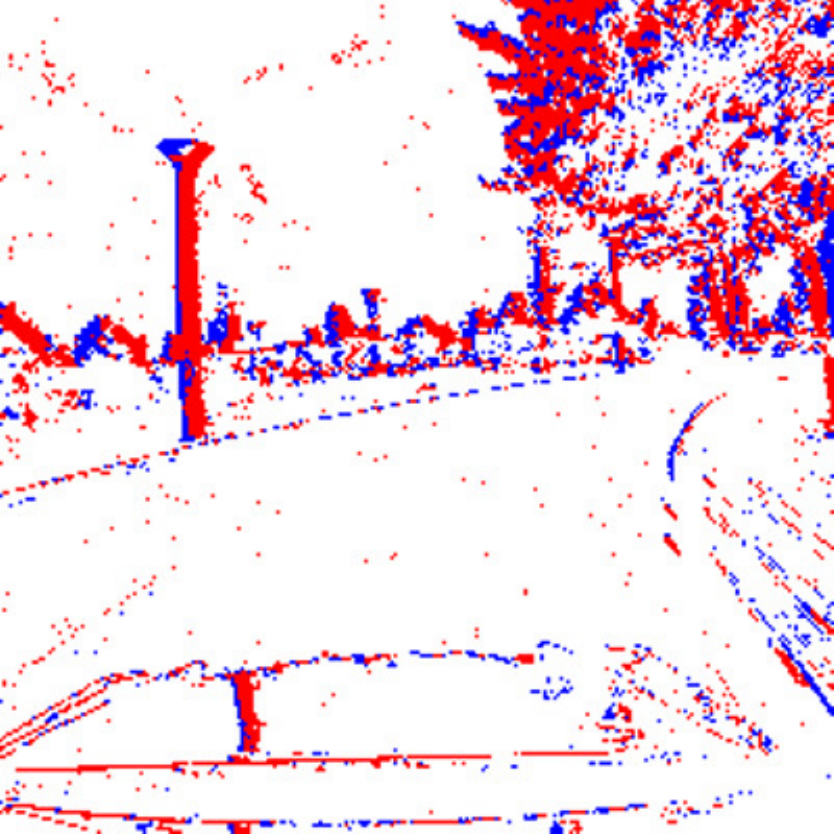}\\
         \vspace{1pt}
        \includegraphics[scale=0.135]{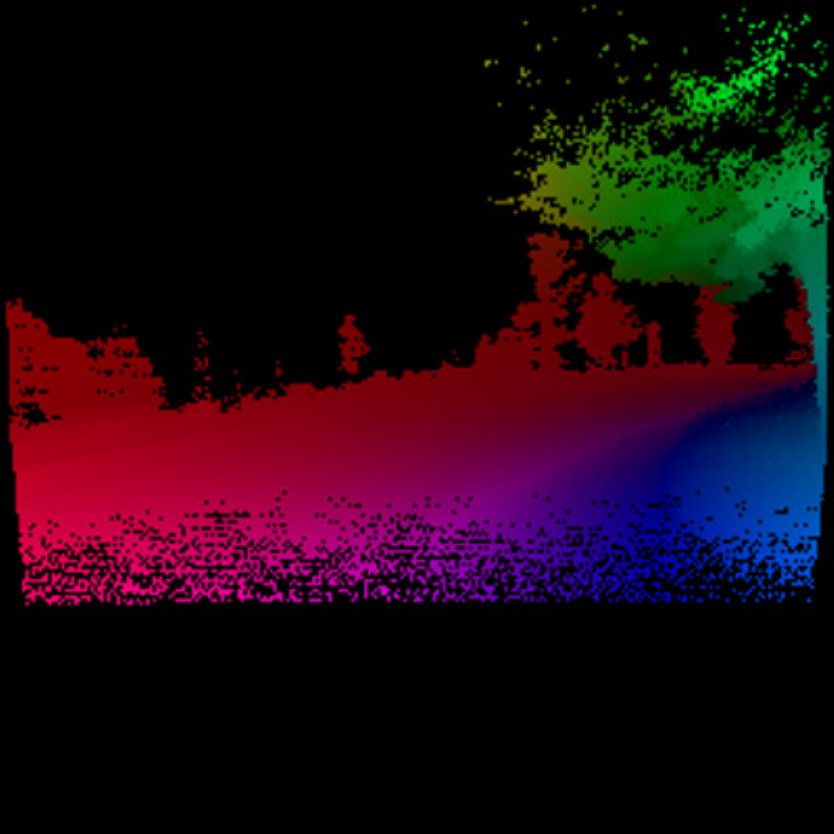}\\
         \vspace{1pt}
         \includegraphics[scale=0.135]{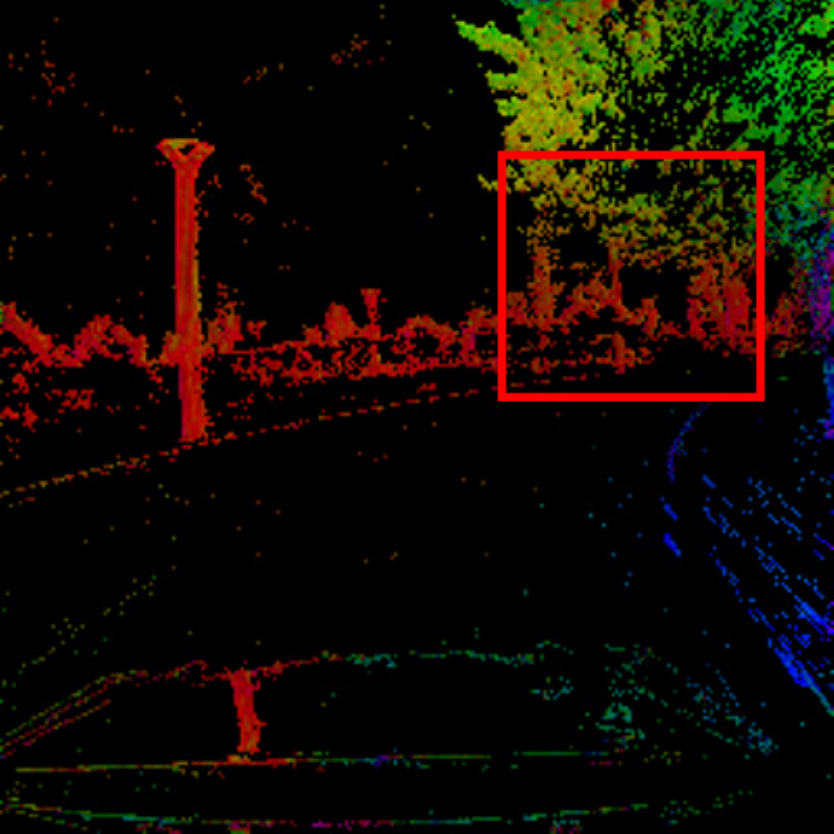}\\
          \vspace{1pt}
         \includegraphics[scale=0.135]{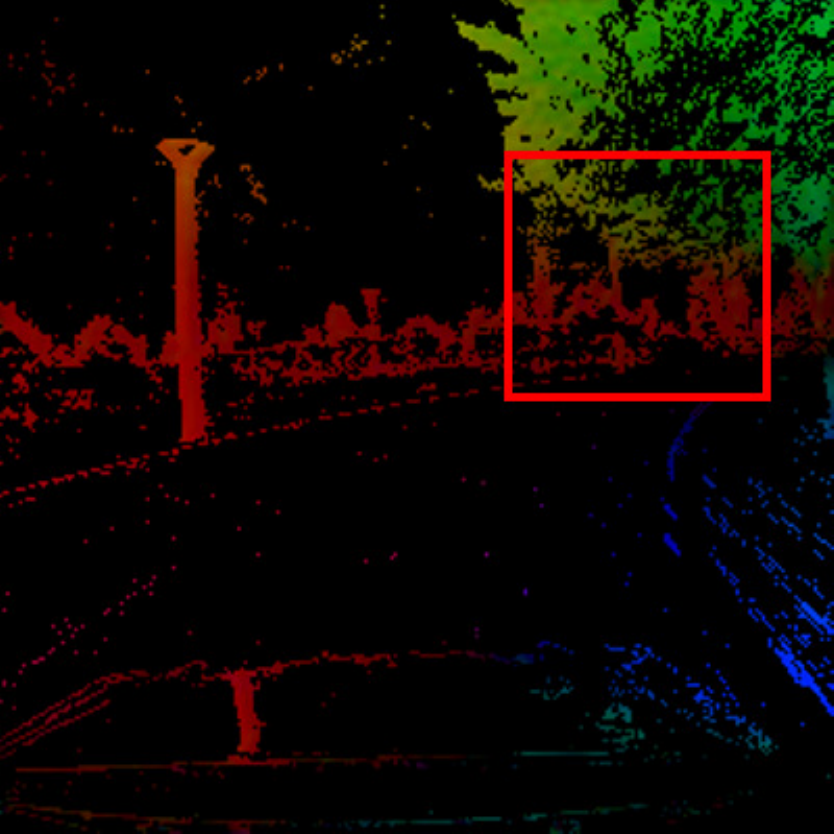}\\
          \vspace{1pt}
         \includegraphics[scale=0.135]{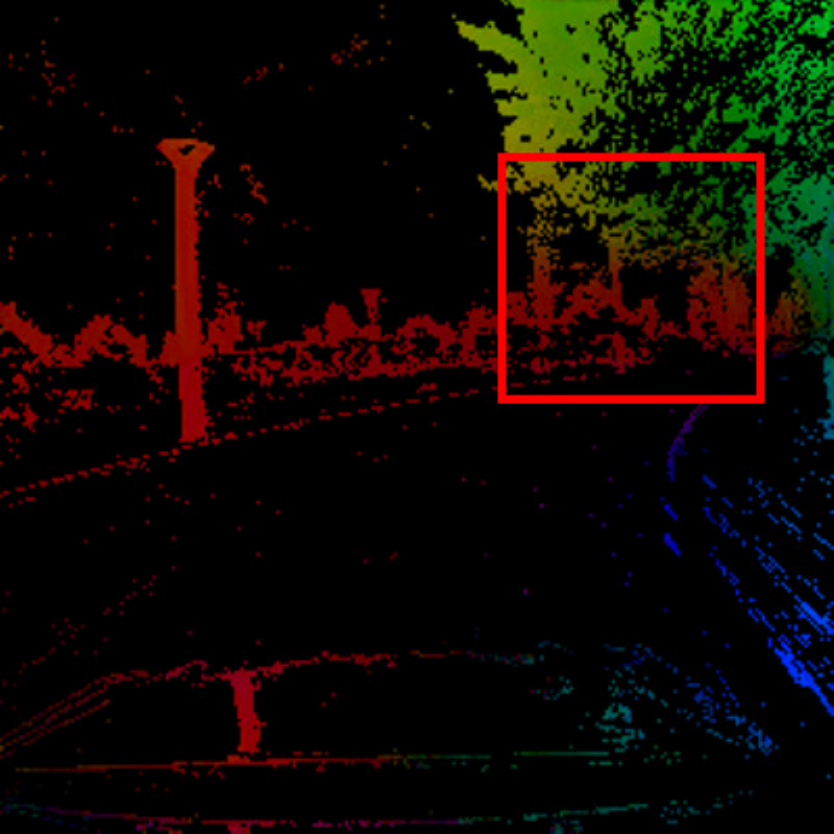}\\
    \end{minipage}
}
 \vspace{-16pt}
\caption{\textbf{Qualitative comparison (dt=4) on MVSEC.} From left to right: indoor\_flying1, indoor\_flying2, indoor\_flying3, and outdoor\_day1.}
\vspace{-0.7cm}
\label{FigureMVSEC}
\end{figure}

\textbf{Metrics.} The metrics for evaluation of prediction accuracy include average endpoint error (AEE), angular error (AE), and the percentage of pixels with AEE greater than 3 pixels (denoted as ``\%Out") for evaluating the robustness to large displacement motions. 

\textbf{Implementation Details.} We train our model implemented with PyTorch on an NVIDIA RTX 2080Ti GPU. The training sequences are split into multiple 128 × 128 (randomly cropped) sequences, each containing 500k events. We use the Adam optimizer with a learning rate of 1e-5 and a batch size of 6. We set the saturation factor $\alpha=0.6$ for $\mathcal{L}_{\mathrm{EC}}$, and the weights $\lambda_1$ and $\lambda_2$ are set to 1 and 0.001.
\subsection{Evaluations on MVSEC and DSEC-Flow}
\begin{figure}[t]
\centering
\includegraphics[scale=0.04]{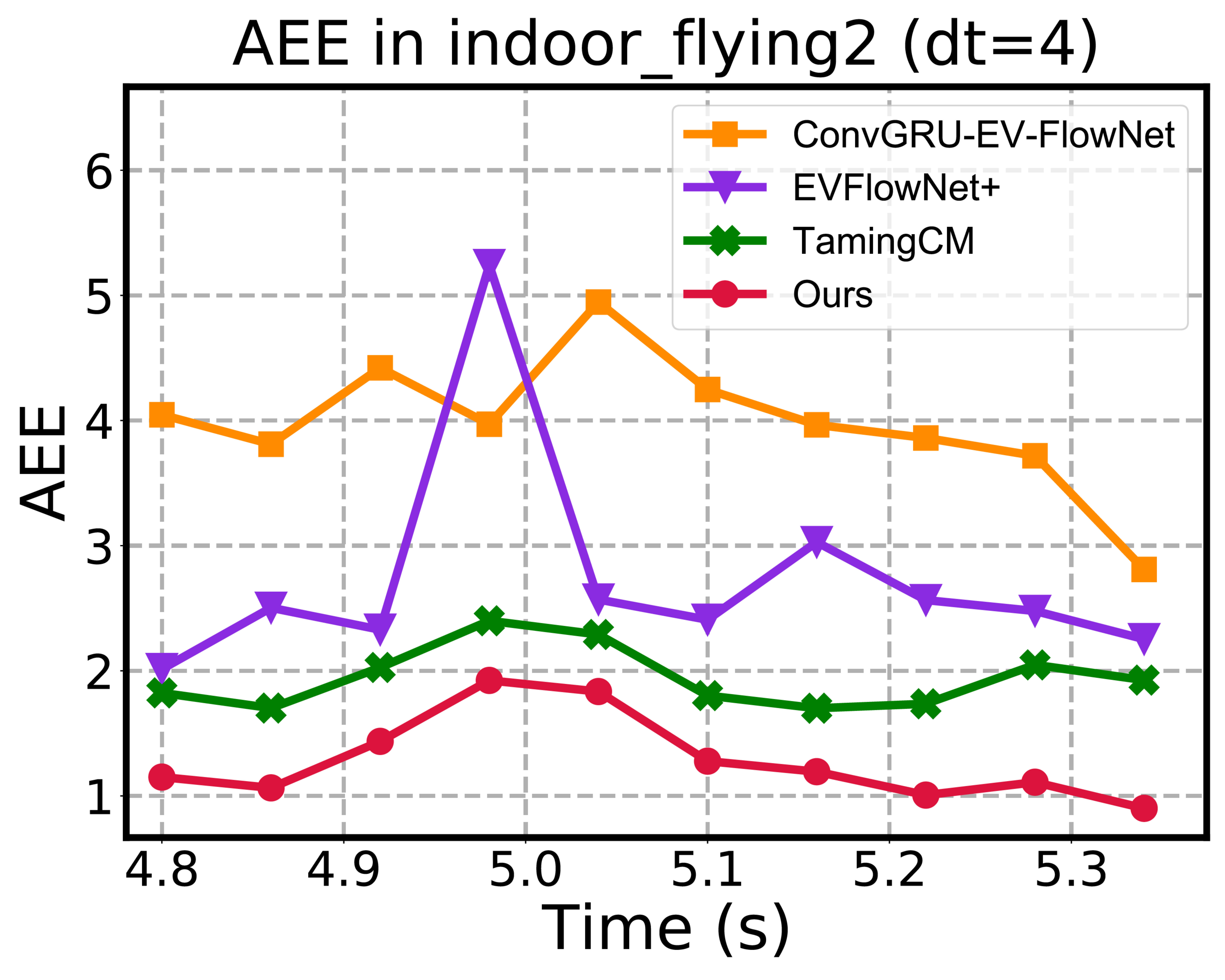}
\includegraphics[scale=0.04]{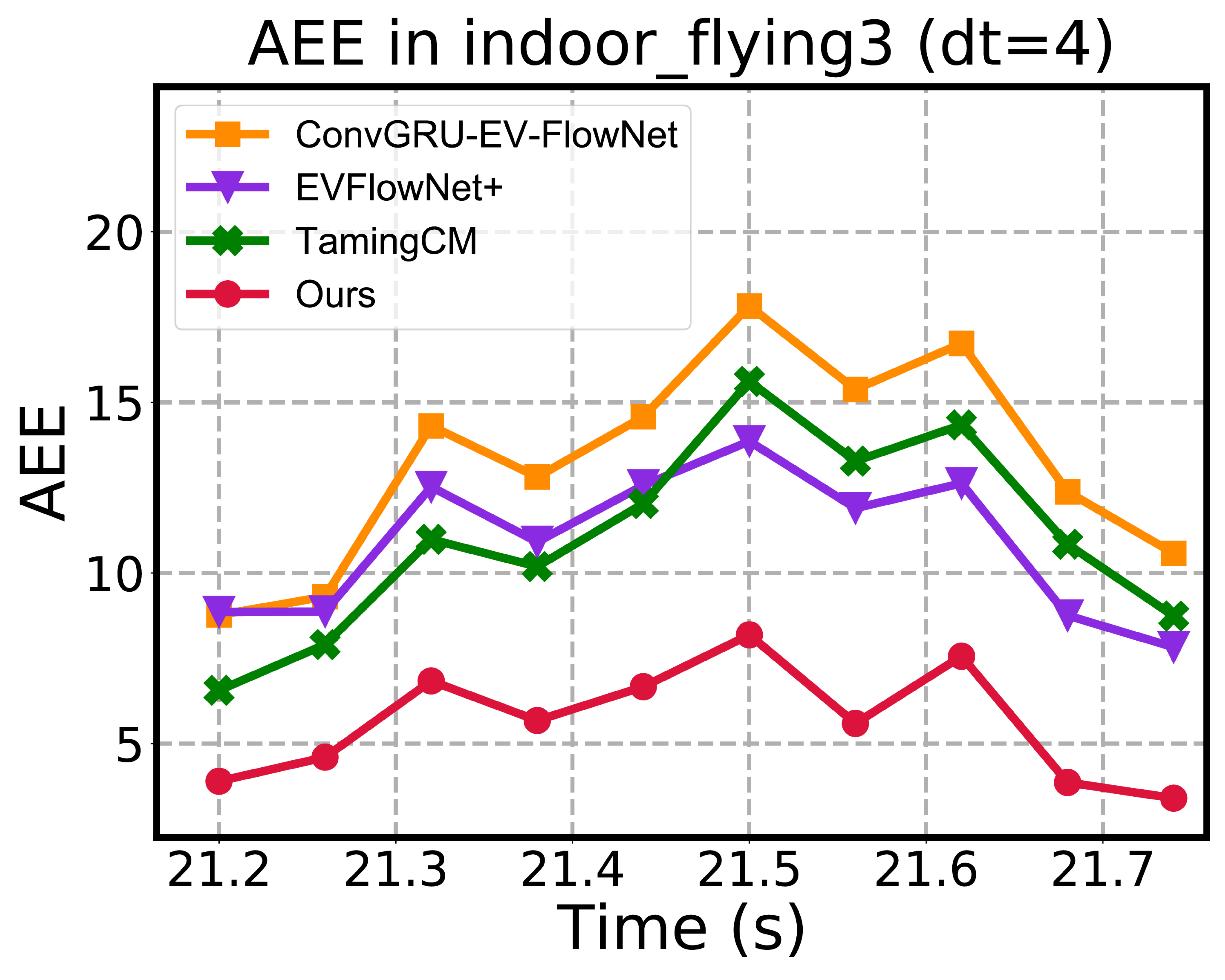}
\caption{\textbf{Performance comparison of state-of-the-art unsupervised methods in challenging scenes (sudden turns).}}
\label{FigureError}
\vspace{-0.65cm}
\end{figure}

\textbf{Evaluation on MVSEC:} Table. \ref{TableMVSEC} reports quantitative results on MVSEC. The top part of the table reports the optical flow corresponding to a time interval of dt=1 grayscale frame (22ms), and the bottom part corresponds to that of dt=4 grayscale frames (89ms). 
Our method outperforms existing unsupervised methods in all indoor sequences and achieves comparable performance in outdoor\_day1 sequence. Specifically, the results for dt=4 reflect the performance of our network in longer-time sequences, which fully proves the effectiveness of our method. Compared with the recent state-of-the-art unsupervised method TamingCM\cite{paredes2023taming}, our method for dt=4 reduces AEE by 19.5\% and \%Out by 37.8\%, respectively. We also give qualitative results for dt=4 as shown in Fig. \ref{FigureMVSEC}, where we compare our method with state-of-the-art unsupervised methods. Consistent with previous works\cite{zhu2019unsupervised,hagenaars2021self,tian2022event,shiba2022secrets,ding2022spatio,zhuang2024ev}, we mask the optical flow predictions with the input events for visualization. 
In addition, to further demonstrate the excellent performance of our method, we compare the performance in challenging scenes (sudden turns of the drone) from indoor flying sequences (MVSEC) with other methods, as shown in Fig. \ref{FigureError}. Specifically, a lower AEE indicates that the estimated optical flow is more accurate. 
The results show that we achieve better performance in challenging scenes, proving the superiority of our method.
\vspace{-0.15cm}

\begin{figure}[t]
\centering
\hspace{-8pt}
\subfigure[]{%
\rotatebox{90}{~~~~~~~~~~~~~~~~~~~~~~~~~~~~~~~~~~\footnotesize ConvGRU-}
\rotatebox{90}{~~~~~\footnotesize \text{Ours} ~~~~~~~~~~\footnotesize TamingCM ~~~~~\footnotesize EV-FlowNet ~~~~~~~~~\footnotesize Events ~~~~~~~~~~~\footnotesize Image}
\begin{minipage}[b]{.26\linewidth}
        \centering
\includegraphics[scale=0.154]{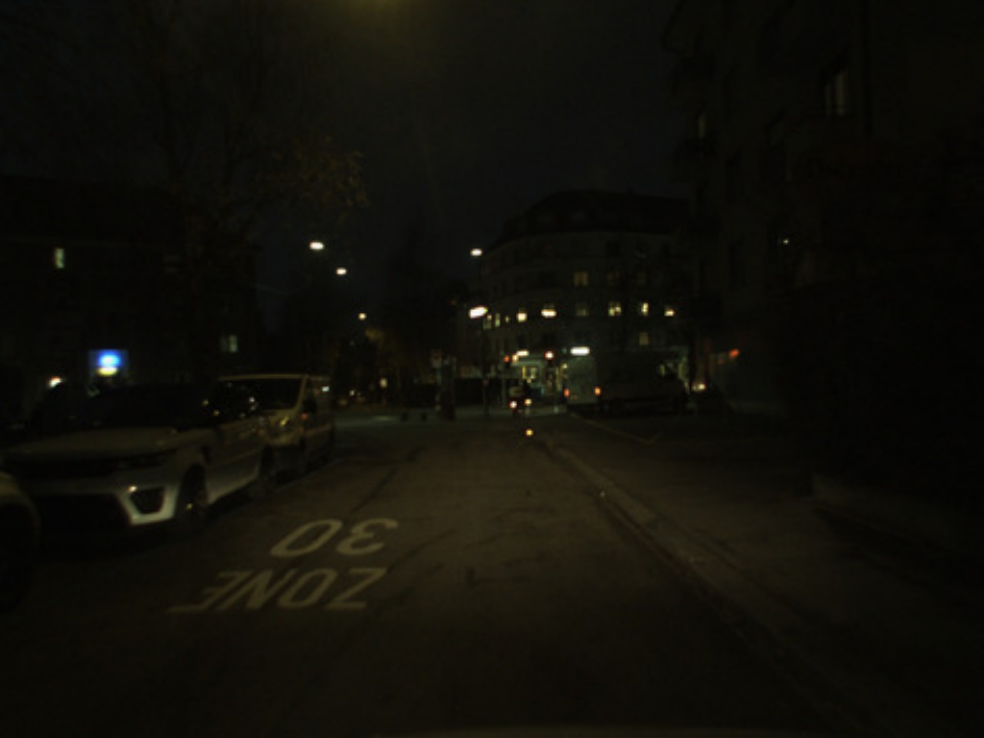}\\
\vspace{1pt}
\includegraphics[scale=0.154]{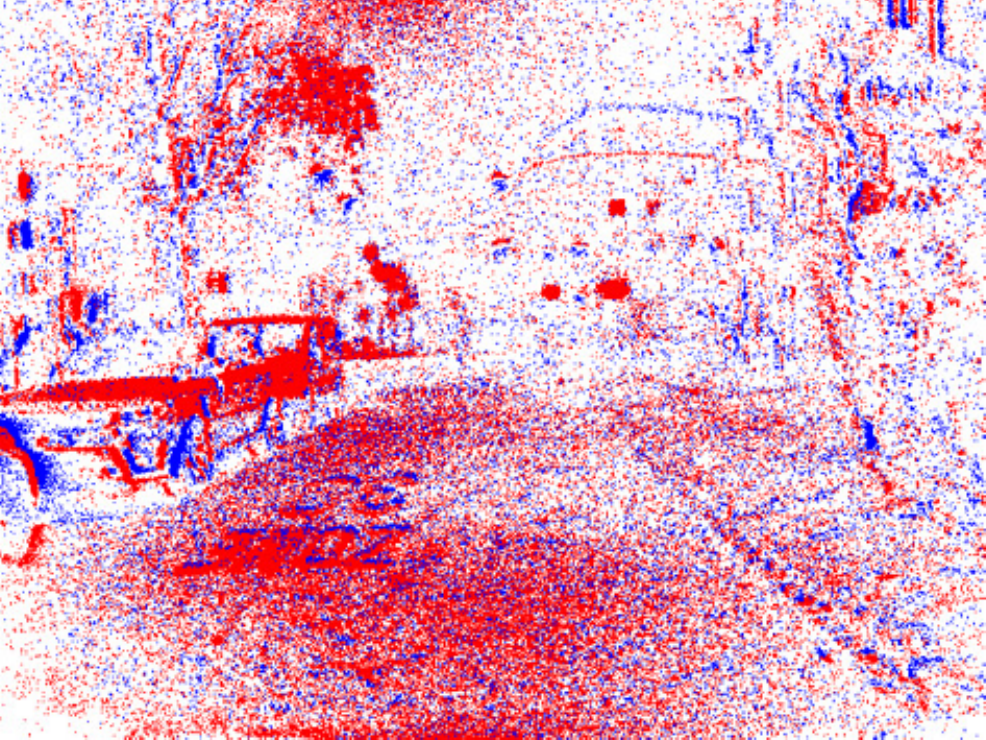}\\
 \vspace{1pt}
\includegraphics[scale=0.154]{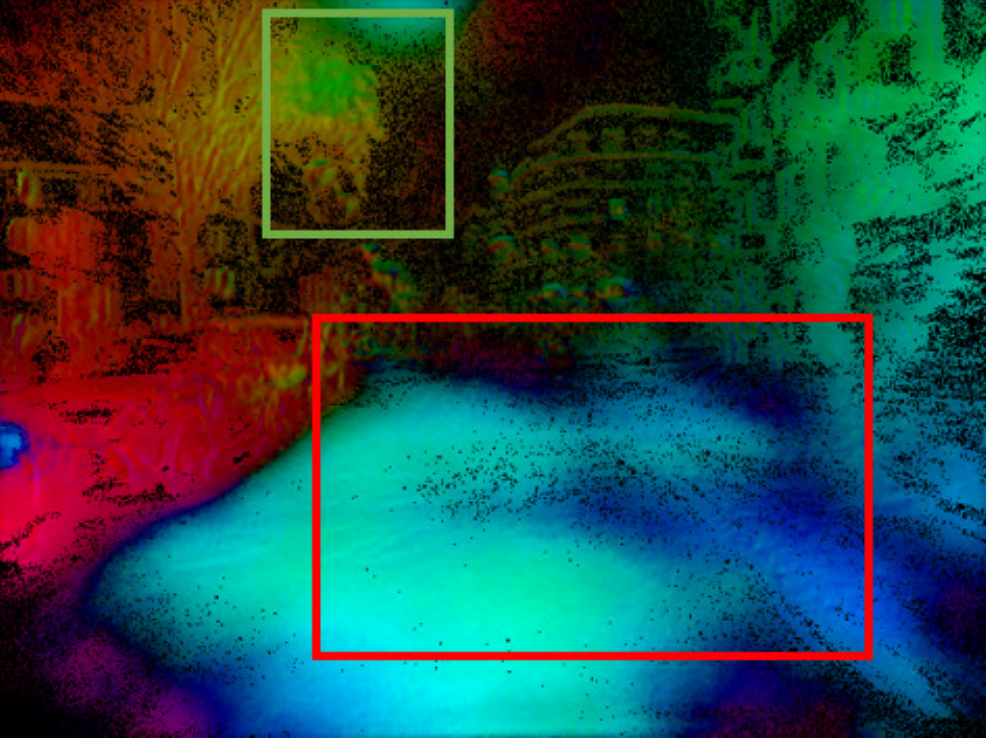}\\
 \vspace{1pt}
\includegraphics[scale=0.154]{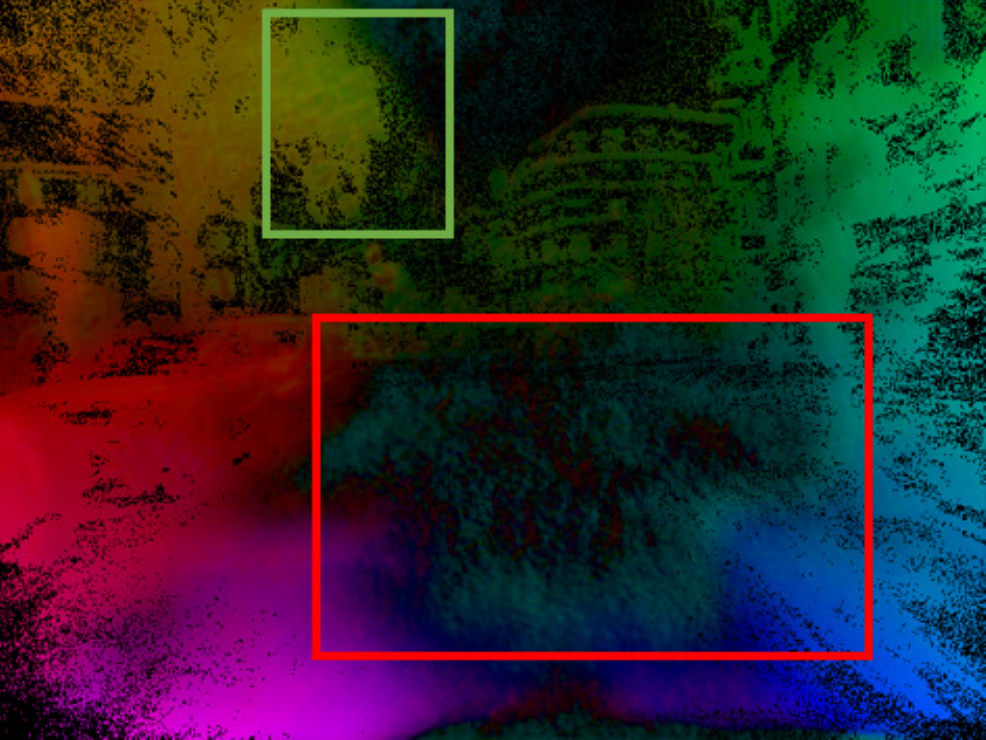}\\
 \vspace{1pt}
\includegraphics[scale=0.154]{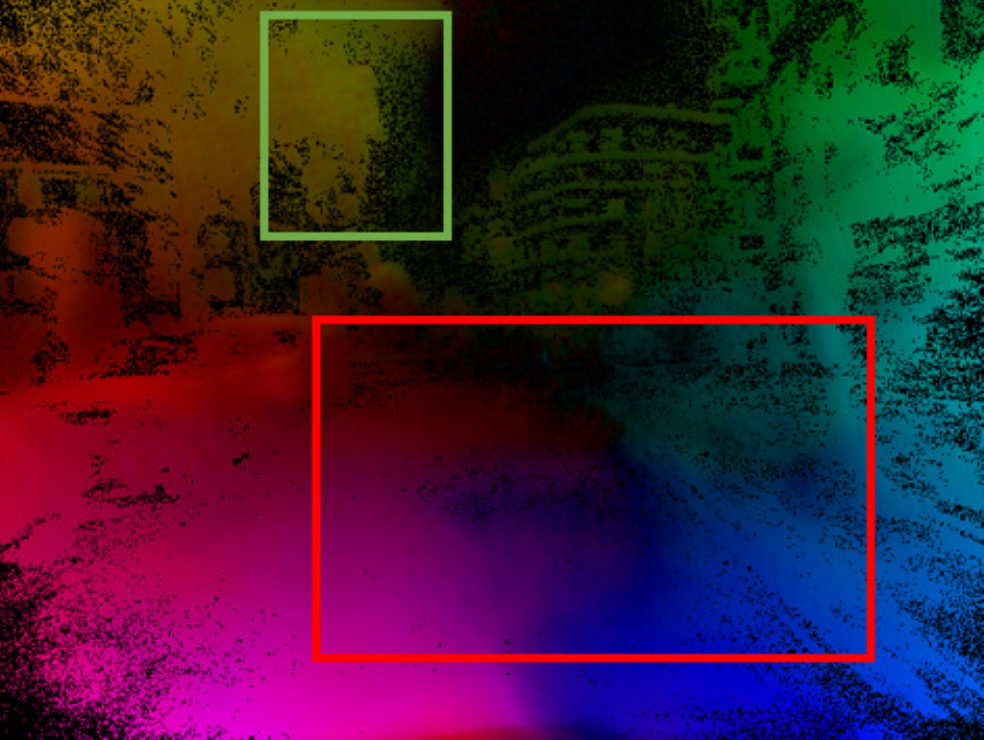}\\
\end{minipage}
}
\subfigure[]{
    \begin{minipage}[b]{.26\linewidth}
        \centering
         \includegraphics[scale=0.154]{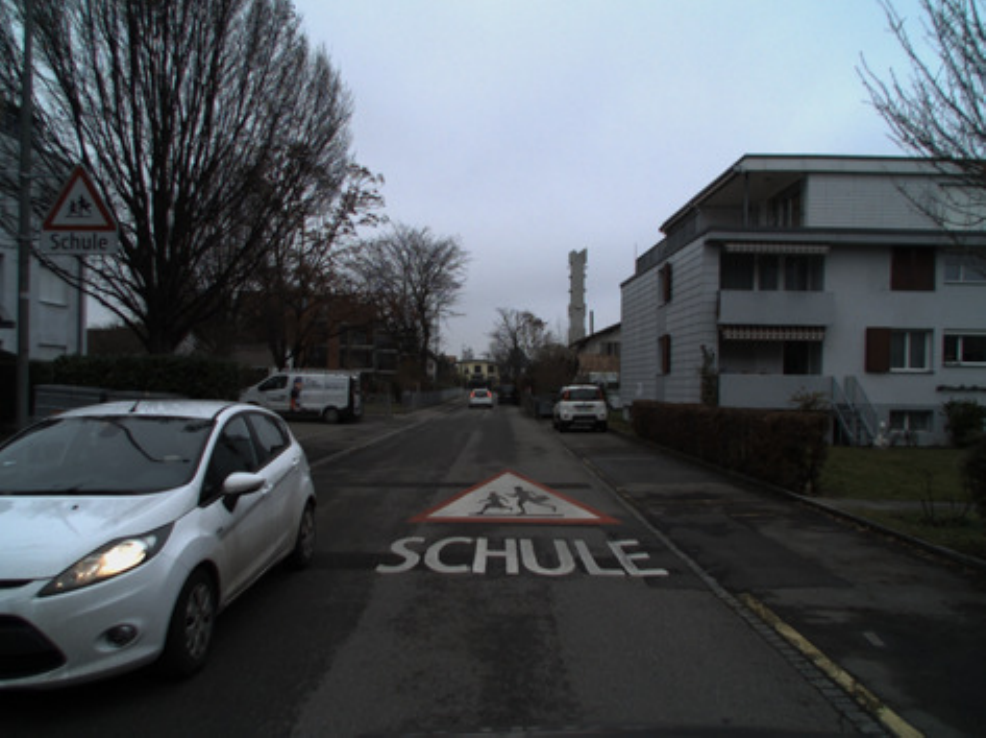}\\
        \vspace{1pt}
         \includegraphics[scale=0.154]{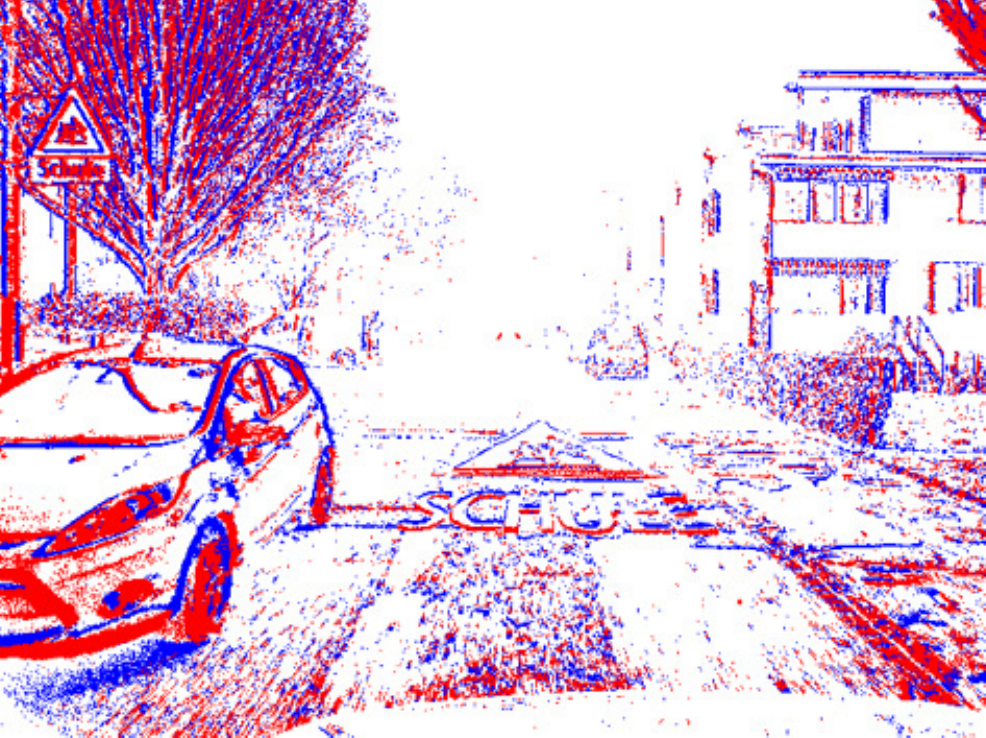}\\
        \vspace{1pt}
        \includegraphics[scale=0.154]{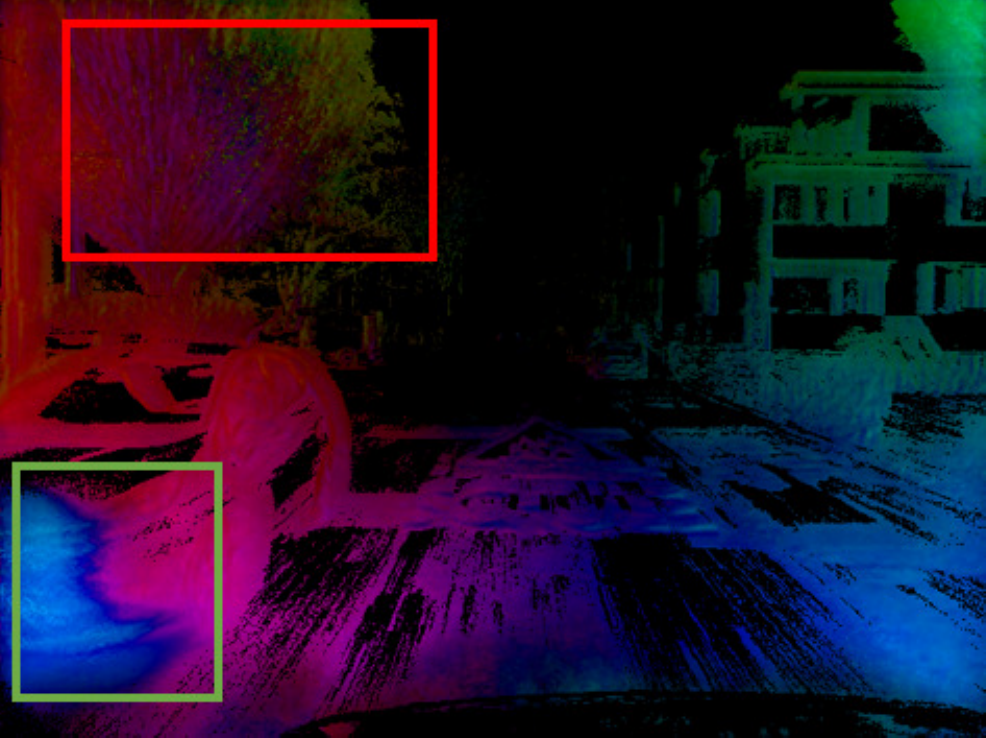}\\
         \vspace{1pt}
        \includegraphics[scale=0.154]{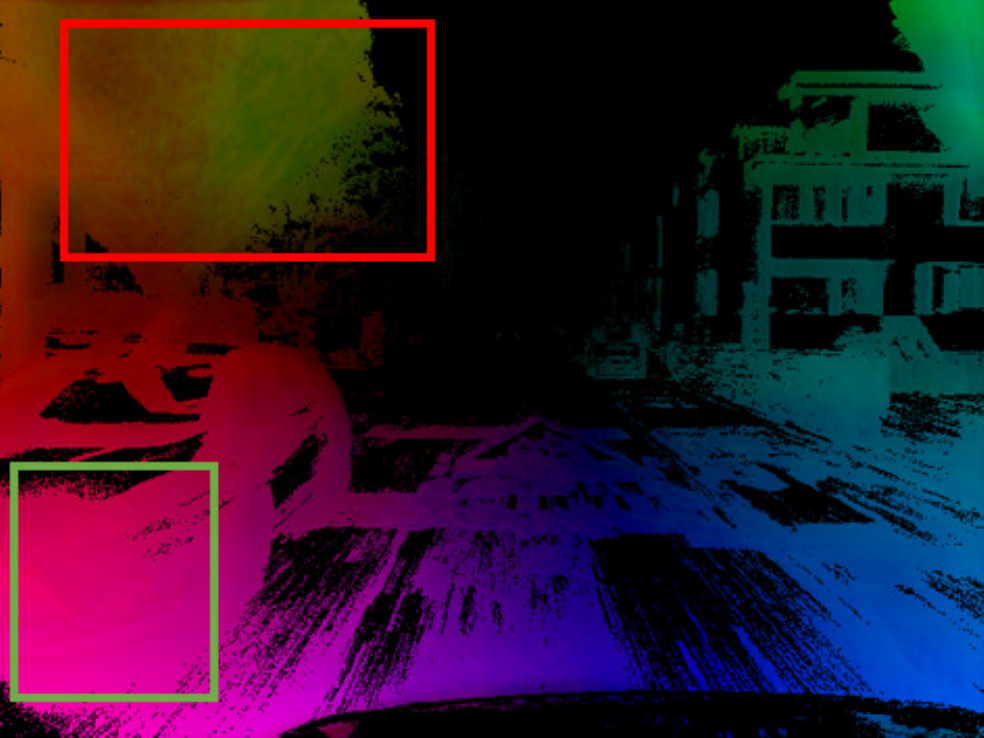}\\
         \vspace{1pt}
         \includegraphics[scale=0.154]{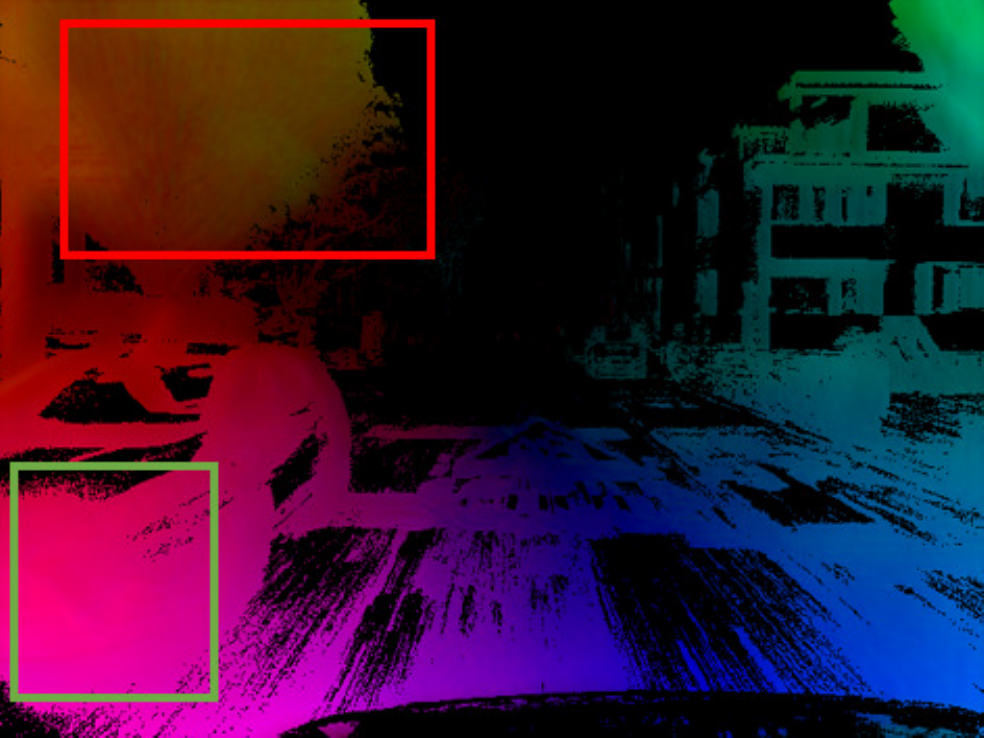}\\
    \end{minipage}
}
\subfigure[]{
    \begin{minipage}[b]{.26\linewidth}
        \centering
         \includegraphics[scale=0.154]{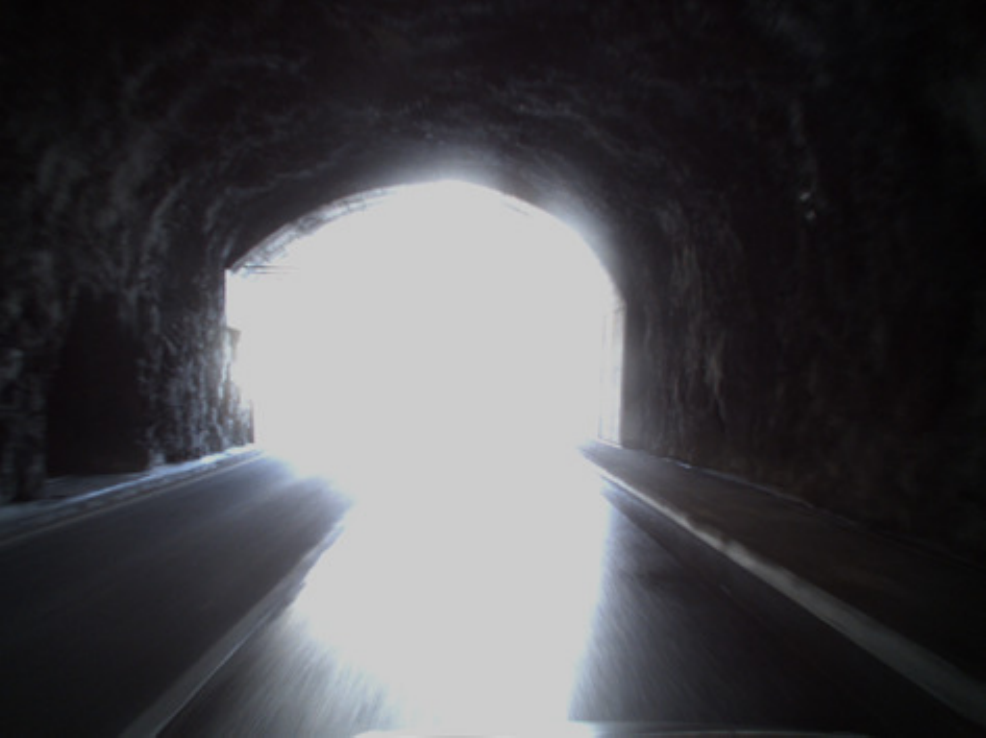}\\
        \vspace{1pt}
         \includegraphics[scale=0.154]{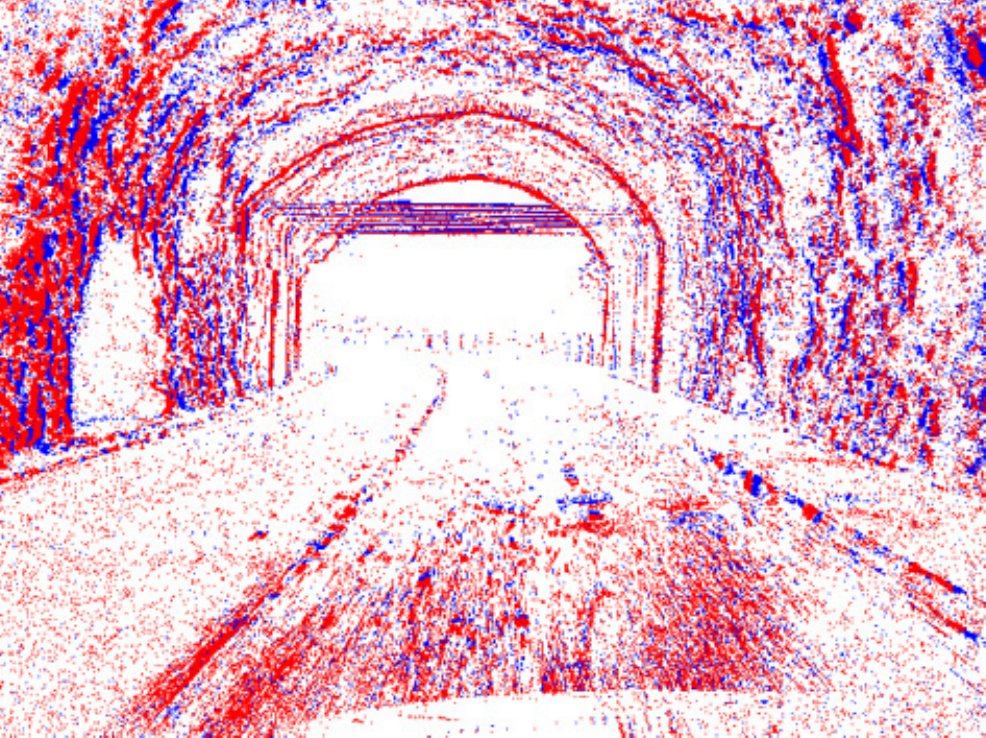}\\
        \vspace{1pt}
        \includegraphics[scale=0.154]{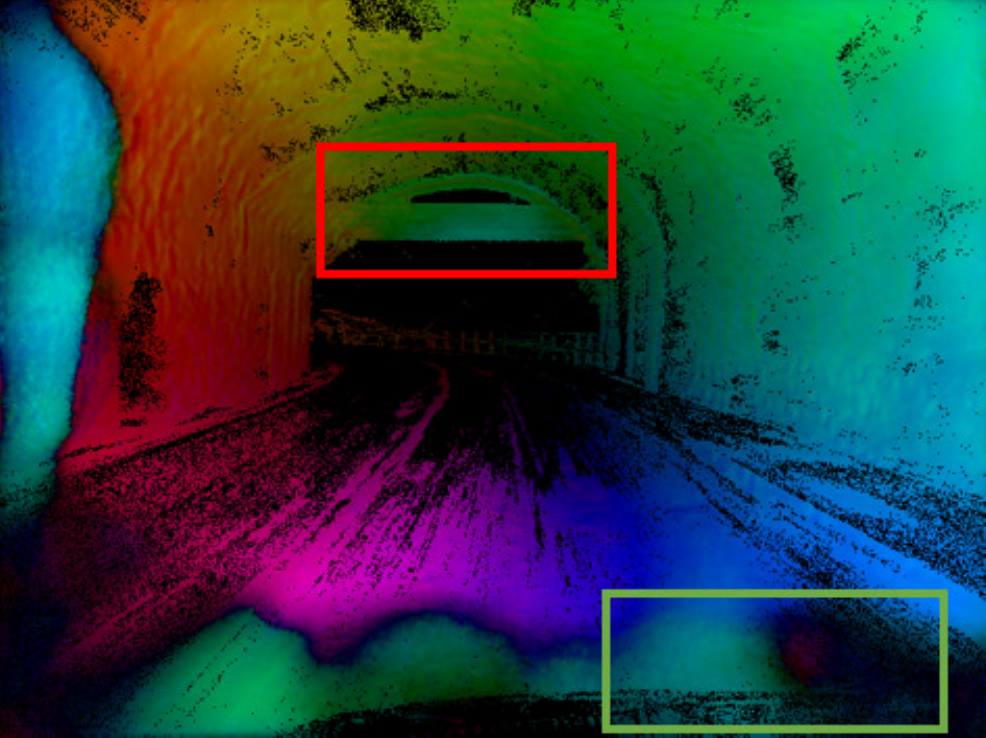}\\
         \vspace{1pt}
        \includegraphics[scale=0.154]{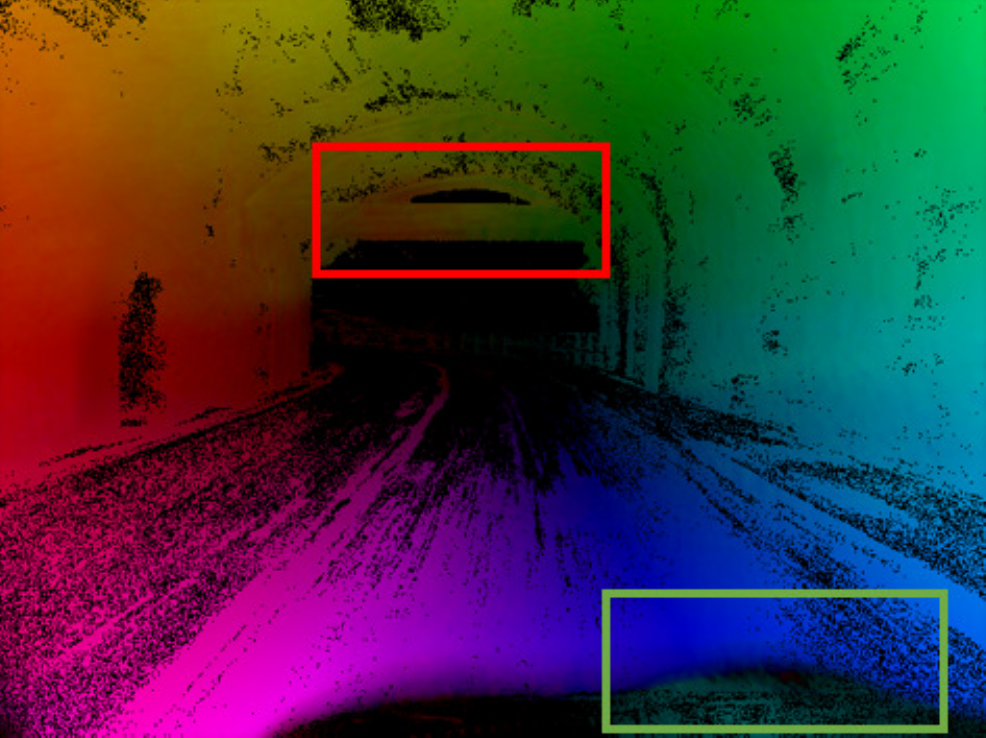}\\
         \vspace{1pt}
         \includegraphics[scale=0.154]{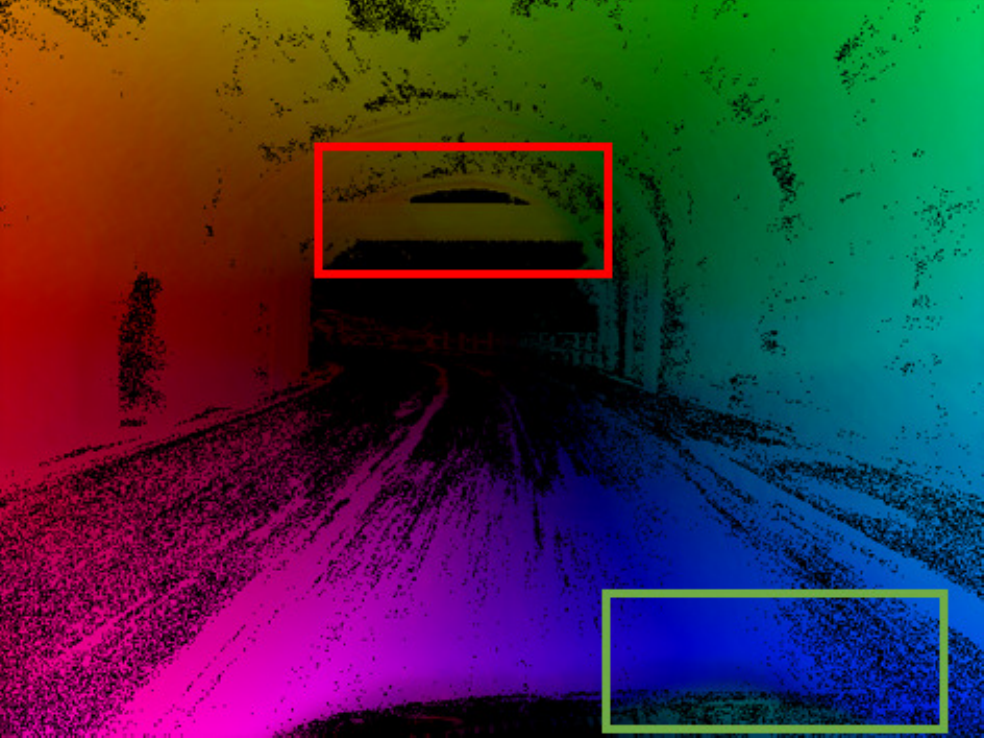}\\
    \end{minipage}
}
 \vspace{-16pt}
\caption{\textbf{Qualitative comparison on DSEC-Flow.} Images are for visualization only, and ground truth is not included due to unavailability. From left to right: zurich\_city\_12\_a, thun\_01\_b and interlaken\_00\_b.}
\vspace{-0.1cm}
\label{FigureDSEC}
\end{figure}

\vspace{-0.1cm}
\textbf{Evaluation on DSEC-Flow:} Our quantitative results on the DSEC-Flow benchmark are presented in Table. \ref{TableDSEC}. The frame interval for optical flow on the DSEC-Flow is 100ms, which is longer than that on MVSEC (22ms). Moreover, DSEC-Flow contains many challenging scenes (such as high-speed motion and high dynamic range). Therefore, the results on DSEC-Flow can further prove the effectiveness of our method in long sequence modeling and the robustness in complex scenes. To demonstrate the superiority of our method as an unsupervised learning approach, we also include the state-of-the-art supervised learning method E-RAFT\cite{gehrig2021raft}. We achieve the best results among all the model-based and unsupervised methods. Compared with TamingCM\cite{paredes2023taming}, our method improves AEE from 2.33 to 2.02 and AE from 17.77 to 13.39. In addition, our method improves \%Out from 10.56 to 9.58, which further demonstrates the robustness of our method in challenging scenes. To prove the superiority of our method in long-time sequences, we accumulate the average estimated optical flow within a time window of 0.1s. The accumulated optical flow results on DSEC-Flow are shown in Fig. \ref{FigureDSEC}. Our results appear smoother and more accurate near image boundaries, which prove the superiority of our method in long-time sequences.

Leveraging the advantages of our method in exploiting rich spatio-temporal motion cues and nonlinear motion between events, we achieve SOTA performance among unsupervised methods, which confirms the effectiveness of our method.
\begin{table}[!t]
\centering
\caption{\textbf{Results on DSEC-Flow.} SL refers to supervised learning methods, MB to model-based methods, and USL to unsupervised learning methods. \textbf{Bold} is best and \underline{underline} is second best.}
\label{TableDSEC}
\resizebox{\linewidth}{!}{
\begin{tabular}{c l c c c}
\toprule[0.05cm]
~ & Methods & AEE↓& AE↓ & \%Out↓ \\ 
\hline\rule{0pt}{8pt}
SL & E-RAFT\cite{gehrig2021raft} & 0.79 & 2.85 & 2.68  \\  
\hline\rule{0pt}{8pt}
\multirow{2}{*}{MB}& Shiba et al.\cite{shiba2022secrets} & 3.47 & 13.98 & 30.86 \\
~ & RTEF\cite{brebion2021real} & 4.88 & 10.82 & 41.95  \\
\hline\rule{0pt}{8pt}
~ & Shiba et al.\cite{shiba2022secrets} & 3.69 & 12.62 & 34.62 \\
~ & ConvGRU-EV-FlowNet\cite{hagenaars2021self} & 4.27 & - & 33.27 \\ 
USL & EV-FlowNet+\cite{zhu2019unsupervised} & 3.86 & - & 31.45 \\
~ & TamingCM\cite{paredes2023taming} & \underline{2.33} & \underline{10.56} & \underline{17.77} \\ 
~ & Ours & \textbf{2.02} & \textbf{9.58} & \textbf{13.39} \\ 
\toprule[0.05cm]
\end{tabular}
}
\vspace{-0.5cm}
\end{table}

\subsection{Ablation Studies}
To demonstrate the effectiveness of different components of our proposed method, we conduct a series of ablation studies on the MVSEC dataset with dt=1 and dt=4 cases. The ablation results are given in Table. \ref{TableAblation}.
We use the recurrent version of EV-FlowNet trained with their compensation loss\cite{hagenaars2021self} as the baseline method (denoted as ``B"). We employ ``N" to denote the proposed nonlinear motion compensation loss (NLMC), while ``A" and ``S" mean the  STMFA module, respectively. 
\begin{table}[!t]
\renewcommand{\arraystretch}{1.0}
\setlength{\tabcolsep}{1.5pt}
    \centering
    \caption{\textbf{Ablation studies of components for dt=1 and dt=4 cases.} ``B" is the baseline method, ``N" means NLMC Loss. ``A" and ``S" refer to AMFE and STMFA module.}
\label{TableAblation}
\resizebox{\linewidth}{!}{
\begin{tabular}{c c c c c c c c c c c c c}
\toprule[0.05cm]
\multirow{2}{*}{dt=1}  & \multirow{2}{*}{B}& \multirow{2}{*}{N}& \multirow{2}{*}{A}& \multirow{2}{*}{S}& \multicolumn{2}{c}{indoor1} & \multicolumn{2}{c}{indoor2} & \multicolumn{2}{c}{indoor3} & \multicolumn{2}{c}{outdoor1} \\ 
\cline{6-13}\rule{0pt}{8pt}
~ &~& ~ & ~ & ~ &  AEE↓ & \%Out↓ & AEE↓ & \%Out↓ & AEE↓ & \%Out↓ & AEE↓ & \%Out↓ \\ 
\hline\rule{0pt}{8pt}
A1 & \checkmark & ~ & ~ & ~ & 0.57 & 0.35 & 1.16  & 8.10 & 0.90 & 5.03  & 0.45 & 0.17 \\ 
A2 &\checkmark& \checkmark & ~ & ~ &  0.42 & 0.24 & 0.89  & 5.13 & 0.69 & 2.86  & 0.32 & 0.08 \\
A3 &\checkmark& \checkmark & \checkmark & ~ &  0.43 & 0.31 & 0.81  & 3.86 & 0.65 & 2.34  & 0.31 & 0.07 \\ 
A4 &\checkmark& \checkmark & ~ & \checkmark &  {0.40} & {0.17} & {0.78} & {3.60} & {0.61} & {1.88} & {0.29} & {0.03} \\
A5 &\checkmark& \checkmark & \checkmark & \checkmark & \textbf{0.39} & \textbf{0.13} & \textbf{0.68}& \textbf{2.23} & \textbf{0.56} & \textbf{1.22} & \textbf{0.27} & \textbf{0.02} \\ 
\hline\rule{0pt}{8pt}
dt=4 \\
\hline\rule{0pt}{8pt}
A1 & \checkmark & ~ & ~ & ~ & 2.09 & 20.95 & 3.90 & 41.54 & 3.00 & 31.65 & 1.65 & 11.72 \\ 
A2 &\checkmark& \checkmark & ~ & ~ &  1.50 & 8.97 & 2.90 & 27.26 & 2.16 & 18.44 & 1.21 & 6.63 \\
A3 &\checkmark& \checkmark & \checkmark & ~ &1.55 & 9.90 & 2.77 & 26.48 & 2.15 & 17.78 & 1.20 & 6.50 \\ 
A4 &\checkmark& \checkmark & ~ & \checkmark & 1.43 & 7.47 & 2.51 & 22.07 & 2.00 & 16.15 & 1.10 & 5.36 \\ 
A5 &\checkmark& \checkmark & \checkmark & \checkmark & \textbf{1.42} & \textbf{6.93} & \textbf{2.26} &\textbf{ 20.09} & \textbf{1.85} & \textbf{13.69}& \textbf{1.04} & \textbf{4.50} \\ 
\toprule[0.05cm]
\end{tabular}
}
\label{table_MAP}
\vspace{-0.55cm}
\end{table}
\vspace{-0.08cm}
Since A1 only considers current motion features and assumes that events move linearly within the loss time window, it performs the worst. Then, compared with A5, removing any modules (in A2$\sim$A4) will lead to performance degradation. Specifically, A2 proves that the nonlinear motion between consecutive events is crucial for optical flow estimation. In addition, as shown in A3 and A4, extracting spatio-temporal information of events effectively and enhancing motion features adaptively can further improve network performance. Especially in A4, this result fully proves that the STMFA module can effectively aggregate the spatio-temporal information of events to estimate accurate optical flow. It is worth mentioning that the experiment for dt=4 case represents the optical flow estimation results of our network in longer-time sequences, which fully shows the robustness of our proposed method in long-time sequences.
\vspace{-0.08cm}
\section{Conclusion}
In this work, we introduce E-NMSTFlow, a novel unsupervised network for event-based optical flow estimation. Our method explores rich spatio-temporal motion cues to establish data associations, enhancing optical flow performance. Meanwhile, we utilize accurate nonlinear motion between events to improve the unsupervised learning of our network. Extensive experiments demonstrate the effectiveness and superiority of our method, and we achieve competitive performance on the MVSEC and DSEC-Flow datasets.
\bibliographystyle{IEEEtran}
\bibliography{main}
\end{document}